\documentclass[lettersize,journal]{IEEEtran}
\usepackage{cite}
\usepackage{amsmath,amssymb,amsfonts}
\usepackage{graphicx}
\usepackage{textcomp}
\usepackage[table,dvipsnames]{xcolor}
\usepackage{graphics}
\usepackage{adjustbox}
\usepackage{graphicx}
\usepackage{threeparttable}
\usepackage{multirow} 
\usepackage{layouts}
\usepackage{enumitem}
\usepackage{amsmath,amsfonts}
\usepackage{array}
\usepackage[caption=false,font=normalsize,labelfont=sf,textfont=sf]{subfig}
\usepackage{textcomp}
\usepackage{stfloats}
\usepackage{url}
\usepackage{verbatim}
\usepackage{graphicx}
\def\BibTeX{{\rm B\kern-.05em{\sc i\kern-.025em b}\kern-.08em
    T\kern-.1667em\lower.7ex\hbox{E}\kern-.125emX}}
\usepackage{balance}

\usepackage{algorithm}
\usepackage{algpseudocode}

\usepackage{fontawesome}
\usepackage{amsfonts}
\usepackage{pifont}
\newcommand{\bata}{\color{LimeGreen!40!black} \faBattery[1] }%
\newcommand{\batb}{\color{LimeGreen!40!black} \faBattery[2]  }%
\newcommand{\batd}{\color{LimeGreen!40!black} \faBattery[4]  }%

\colorlet{mc}{LimeGreen!50!White!50!}
\newcommand{\hlc}{\cellcolor{mc}}

\newcommand{\mytilde}{\raisebox{0.5ex}{\texttildelow}}

\usepackage{xspace}
\newcommand{\axname}{\mathrm{ml_{cax}}}
\newcolumntype{?}{!{\vrule width 1pt}}
\makeatletter
\newcommand{\thickhline}{%
    \noalign {\ifnum 0=`}\fi \hrule height 1pt
    \futurelet \reserved@a \@xhline
}
\newcolumntype{"}{@{\hskip\tabcolsep\vrule width 1pt\hskip\tabcolsep}}
\makeatother

\begin{document}

%
\title{Model-to-Circuit Cross-Approximation \\For Printed Machine Learning Classifiers}

%
%
%
%

\author{Giorgos Armeniakos,
        Georgios Zervakis,
        Dimitrios Soudris,~\IEEEmembership{Member,~IEEE,}\\
        Mehdi B. Tahoori,~\IEEEmembership{Fellow,~IEEE},
        and J{\"o}rg Henkel,~\IEEEmembership{Fellow,~IEEE}
        
\thanks{Manuscript received July 1, 2022, revised January 10, 2023. \textit{Corresponding Author: Giorgos Armeniakos (georgios.armeniakos@kit.edu)}}%
\thanks{G.~Armeniakos and D.~Soudris are with the School of Electrical and Computer Engineering, National Technical University of Athens, Athens 15780, Greece. G.~Armeniakos was also with the Karlsruhe Institute of Technology (KIT), Karlsruhe 76131, Germany.}%
\thanks{G. Zervakis is with the Computer Engineering \& Informatics Dept., University of Patras, Patras 26504, Greece. This research was done when he was with the Karlsruhe Institute of Technology (KIT).}
\thanks{M.~B.~Tahoori and J.~Henkel are with the Department of Computer Science, Karlsruhe Institute of Technology (KIT), Karlsruhe 76131, Germany.}%
\thanks{This work is partially supported by the German Research Foundation (DFG) through the project ``ACCROSS: Approximate Computing aCROss the System Stack'' HE 2343/16-1 and by grant from the Excellence Initiative of Karlsruhe Institute of Technology under Future Field program ``SoftNeuro''.}
}

\newcommand{\red}[1]{{\color{red}#1}}
\newcommand{\blue}[1]{{\color{black}#1}}
\newcommand{\orange}[1]{{\color{black}#1}}
\newcommand{\yellow}[1]{{\color{black}#1}}



\maketitle
\begin{abstract}
Printed electronics (PE) promises on-demand fabrication, low non-recurring engineering costs, and sub-cent fabrication costs.
It also allows for high customization that would be infeasible in silicon, and bespoke architectures prevail to improve the efficiency of emerging PE machine learning (ML) applications.
Nevertheless, large feature sizes in PE prohibit the
realization of complex ML models in PE, even with bespoke architectures.
In this work, we present an automated, cross-layer approximation framework tailored to bespoke architectures that enable complex ML models, such as Multi-Layer Perceptrons (MLPs) and Support Vector Machines (SVMs),  in PE.
Our framework adopts cooperatively a hardware-driven coefficient approximation of the ML model at algorithmic level, a netlist pruning at logic level, and a voltage over-scaling at the circuit level.
Extensive experimental evaluation on 12 MLPs and 12 SVMs and more than 6000 approximate and exact designs demonstrates that our model-to-circuit cross-approximation delivers power and area optimal designs that, compared to the state-of-the-art exact designs, feature on average 51\% and 66\% area and power reduction, respectively, for less than 5\% accuracy loss.
Finally, we demonstrate that our framework enables 80\% of the examined classifiers to be battery-powered with almost identical accuracy with the exact designs, paving thus the way towards smart complex printed applications.

\end{abstract}


\section{Introduction}\label{sec:introduction}

\IEEEPARstart{P}{rinted} electronics have attracted a major interest,
mainly due to their prominent characteristics of light-weight, bendable, and low-cost hardware, promising future developments in consumer electronics.
Printed electronics opens the door to a variety of new applications in various sectors like healthcare, disposables~\cite{Bio:healthcare}, as well as in the emerging technology of smart packaging (e.g. packaged foods, beverages), fast moving consumer goods (FMCG), detecting devices and more~\cite{Mubarik:MICRO:2020:printedml}.
The main challenge in such domains is to produce ultra-low cost conformal battery- or even self-powered devices.
Despite the high efficiency of silicon systems, their high manufacturing cost and inadequacy to meet tight  flexibility, stretchability, etc., requirements~\cite{Bleier:ISCA:2020:printedmicro}, puts printed electronics in the spotlight.

\begin{figure*}[t!]
\centering
\includegraphics[]{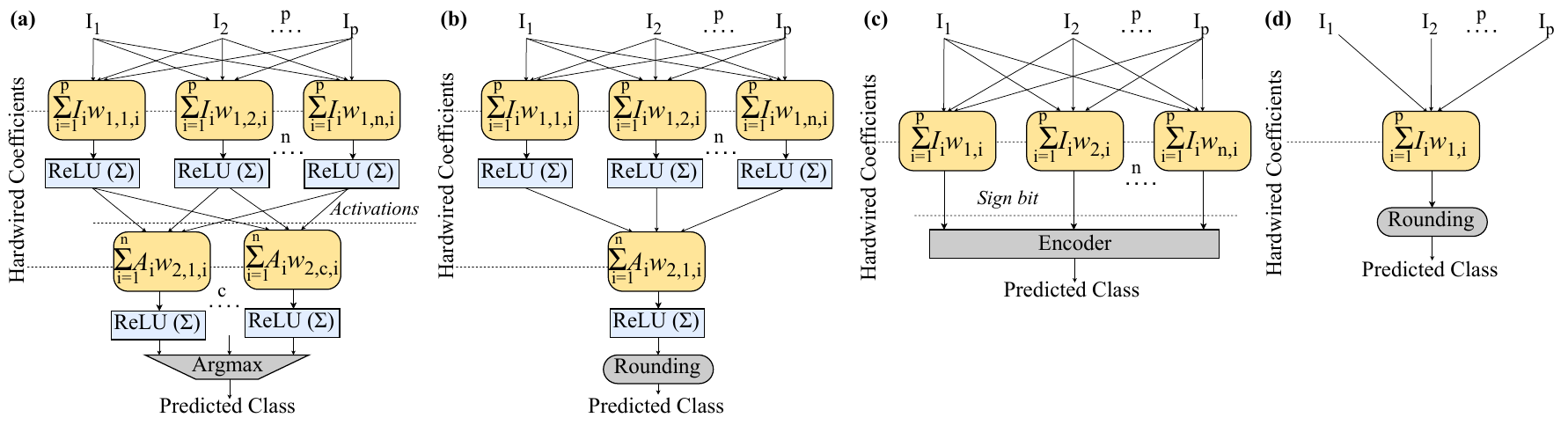}
\caption{Architectures of bespoke ML classifiers. a) MLP-C, b) MLP-R, c) SVM-C, d) SVM-R
}
\label{fig:architectures}
\end{figure*}

The main limitation of printed devices is their extremely large feature sizes.
The associated hardware overheads act prohibitively for most complex circuits, including implementations of Machine Learning (ML) classification algorithms essential for numerous applications in aforementioned domains~\cite{Mubarik:MICRO:2020:printedml}.
Indicatively, it has been demonstrated that a printed  multiply-and-accumulate (MAC) unit (i.e., most common in ML circuits) features \mytilde6 orders of magnitude higher area and $8\times$ higher power than a silicon based one at the 40nm technology node~\cite{Mubarik:MICRO:2020:printedml}. 
Hence,  due to the aforementioned constraints, research activity on printed ML classifiers is still very limited.

To address these restrictions and exploiting the low-cost in-situ fabrication offered by printed electronics, \emph{bespoke circuits} have emerged as a prominent solution~\cite{BespokeProcessor,Mubarik:MICRO:2020:printedml,Bleier:ISCA:2020:printedmicro,Whatmough2019FixyNNEH}.
\yellow{The term \emph{bespoke} refers to fully-customized circuit implementations, even per ML model and dataset.}\label{commentR1C4}
Leveraging the potential for high customizations, ~\cite{BespokeProcessor} designed a bespoke processor that, based on the gate-level activity analysis of a specific application, removes unused gates.
Authors in~\cite{Mubarik:MICRO:2020:printedml} examined printing bespoke ML circuits, in which model parameters are hardwired in the circuit's description, paving the way towards ML classification on printed technologies.
However, due to severe hardware overheads, \cite{Mubarik:MICRO:2020:printedml} deduced that only simple ML algorithms such as Decision Trees and simple Support Vector Machine Regression (SVM-R) are realistic targets for ultra-resource constrained printed circuits.

An efficient solution to reduce the associated complexity of ML circuits is to exploit their intrinsic error resilience and trade computational accuracy for diminished hardware overheads by leveraging Approximate Computing (AC) principles~\cite{Shafique:DAC:2016:cross}.
Designing approximate arithmetic units such as adders and multipliers has attracted a vast research interest and has been demonstrated as a dominant solution in many neural networks accelerators~\cite{survey:armen}.
\orange{
Moreover, there exist significant research interest in gate-level pruning~\cite{GatePrun2017,Scarabottolo:DAC:2019:prune,Zervakis:ACCESS2020}
that act at logic level and have an inherent efficiency in reducing a circuit's area complexity.
Gate-level pruning technique operates on an already optimized netlist and further reduces its area by removing selected gates.}
Another widely used and power-efficient technique at physical (circuit) level is voltage overscaling (VOS)~\cite{vader:zerv,vosim:zervakis,Zervakis2019MultiLevel,zerv:axmult}.
VOS scales the voltage supply value beyond its nominal value, leading though to unsustainable clock frequencies.
Due to the quadratic dependence of supply voltage on dynamic power consumption, VOS delivers significant energy savings, but with the potential of erroneous results due to timing violations~\cite{vader:zerv}. 
Logic approximation and VOS are proven to exhibit a synergistic nature, if they are systematically applied~\cite{Zervakis2019MultiLevel,vader:zerv}.

In our preliminary work~\cite{DATE22:Armen} we combined bespoke implementations with algorithmic and logic approximations investigating, for the first time, approximate printed ML classifiers.
In this work, we extend~\cite{DATE22:Armen} by leveraging the delay slack due to approximation to apply VOS and further boost the power savings at the cost of only a small increase in accuracy loss.
In addition, we present a genetic-based optimization to quickly traverse the newly defined design space and extract approximate printed ML classifiers that satisfy the given battery constraints while achieving relatively high accuracy.
Overall, our framework applies i) ML model approximation by approximating the model's coefficients, i.e., replacing the coefficients of a given trained model with more (bespoke) hardware-friendly values,
ii) logic approximation through a netlist pruning approach customized for printed bespoke architectures,
and iii) circuit-level approximation using VOS.
Using our framework, we elucidate the impact of a holistic cross-layer approximation that is proven to outperform single layer techniques~\cite{Shafique:DAC:2016:cross,Zervakis2019MultiLevel} on designing complex printed ML circuits.
Our extensive evaluation over 6000 approximate ML designs demonstrates that our framework decreases the area and power by 51\% and 66\% on average, respectively, for less than 5\% accuracy loss.
Moreover, our framework allows 80\% of the examined classifiers to be battery-powered and operate with a negligible accuracy loss ($<\!1\%$).
In~\cite{DATE22:Armen} we considered 12 classifiers while in this work we present an extensive and diverse evaluation considering 24 ML classifiers.

\noindent
\textbf{Our novel contributions within this work are as follows}:
\begin{enumerate}[topsep=0pt,leftmargin=*]
    \item This is the first work that evaluates the impact of holistic model-to-circuit cross-approximation in the design and realization of printed ML classifiers. 
    \item We propose an automated framework to generate close-to-accuracy-optimal cross-approximated printed ML circuits under given battery constraints.
    \item We demonstrate, for the first time, that approximate computing enables the realization of complex battery powered printed ML classifiers with minimal accuracy degradation. 
\end{enumerate}

\section{Background}\label{sec:background}
Printed electronics are generated using both subtractive and additive manufacturing processes.
The subtractive ones are similar to the more silicon industry standard subtractive copper etching processes, which involve costly lithography.
However, the additive processes do not involve etching, and hence the production chain is simplified substantially~\cite{chang2014fully} and the cost becomes lower.

\orange{
Printed electronics denotes a set of printing methods which can realize ultra low-cost \cite{subramanian2008printed}, large area \cite{chen201430} and flexible \cite{mohammed2017all} computing systems in which the sensors, actuators, computation and even the energy source are realized using functional materials and inks on the same substrate. There are different processes for the fabrication of printed circuits, such as screen printing,  jet-printing  or roll-to-roll processes \cite{de2010fully}. 

Printed electronics do not compete with silicon-based electronics in terms
of integration density, area and performance. Typical frequencies achieved by
printed circuits are from a few Hz to a few
kHz~\cite{cadilha2017digital}. Similarly, the feature
size tends to be several
microns~\cite{lei2019low}. However, due to its form-factor, conformity and most importantly, significantly reduced fabrication costs -- even for low quantities -- it can target application domains, unreachable by conventional silicon-based VLSI. 

Despite these attractive features, there are several limitations of printed electronics compared to traditional silicon technologies. Due to large feature sizes, the integration density of printed circuits is orders of magnitude lower than silicon VLSI. Additionally, due to large intrinsic transistor gate capacitances, the performance of printed circuits is orders of magnitude lower compared to nanometer technologies. 

}

\begin{table*}[t]
\setlength\tabcolsep{3pt}
\caption{Evaluation of Bespoke Printed ML Circuits in EGT PDK library.}
\label{tab:baselines}
\footnotesize
\centering
\renewcommand{\arraystretch}{1.2}
\begin{threeparttable}
\begin{tabular}{l|cccccc|cccccc|cccccc|cccccc}
\hline
 & \multicolumn{6}{c|}{\textbf{MLP-C}} & \multicolumn{6}{c|}{\textbf{MLP-R}} & \multicolumn{6}{c|}{\textbf{SVM-C}} & \multicolumn{6}{c}{\textbf{SVM-R}} \\ \hline
 & Ac\tnote{1}  & T\tnote{2} & \#C\tnote{3}  & A\tnote{4} & P\tnote{5} & D\tnote{6}
 & Ac\tnote{1}  & T\tnote{2} & \#C\tnote{3}  & A\tnote{4} & P\tnote{5} & D\tnote{6}
 & Ac\tnote{1}  & T\tnote{2} & \#C\tnote{3}  & A\tnote{4} & P\tnote{5} & D\tnote{6}
 & Ac\tnote{1}  & T\tnote{2} & \#C\tnote{3}  & A\tnote{4} & P\tnote{5} & D\tnote{6}
\\ \hline 
\textbf{Cardio}    & 0.88 & (21,3,3)  & 72  & 33.4 & 124.2 & 123 & 0.83 & (21,3,1)  & 66 & 21.6 & 78.1 & 119 & 0.90 & 3  & 63  & 15.1 & 57.4 & 75 & 0.84 & 1  & 21 & 6.8  & 26.6 & 82\\
\textbf{RedWine}   & 0.56 & (11,2,6)  & 34  & 17.6 & 73.5 & 138 & 0.56 & (11,2,1)  & 24 & 7.1  & 28.9 & 101 & 0.57 & 15 & 66  & 23.5 & 92.8 & 66 & 0.56 & 1  & 11 & 4.0  & 18.9 & 77\\
\textbf{WhiteWine} & 0.54 & (11,4,7)  & 72  & 31.2 & 126.4 & 141 & 0.53 & (11,4,1)  & 48 & 13.1 & 48 & 125 & 0.53 & 21 & 77  & 28.3 & 112.4 & 60 & 0.53 & 1  & 11 & 4.2  & 18.9 & 83\\
\textbf{Seeds}    & 0.94 & (7,3,3)  & 30  & 9.9 & 45 & 134 & 0.87 & (7,3,1)  & 25 & 8.3 & 33 & 118 & 0.92 & 3  & 63  & 6.7 & 30.6 & 65 & 0.75 & 7  & 21 & 3.7  & 17.7 & 87\\
\textbf{Vertebral 3C}   & 0.83 & (6,3,3)  & 27  & 8.8 & 41.9 & 116 & 0.72 & (6,3,1)  & 21 & 8.5  & 34.6 & 115 & 0.84 & 3 & 66  & 4.0 & 20.9 & 58 & 0.66 & 1  & 6 & 2.9  & 14.3 & 80\\
\textbf{Balance Scale} & 0.91 & (4,3,3)  & 21  & 9.3 & 39.6 & 117 & 0.86 & (4,3,1)  & 15 & 5.5 & 24.4 & 94 & 0.89 & 3 & 77  & 1.9 & 9.7 & 56 & 0.81 & 1  & 4 & 2.1  & 10.0 & 67\\ \hline
\end{tabular}
\begin{tablenotes}\footnotesize
\item[] $^1$ Accuracy using $8$-bit coefficients and $4$-bit inputs.
$^2$ Model's topology (for SVMs: the number of classifiers).
$^3$ Number of coefficients of the model.
$^4$ Area in $cm^{2}$.
$^5$ Power in $mW$.
$^6$ Delay in $ms$.
\end{tablenotes}
\end{threeparttable}
\end{table*}

\section{Bespoke Machine Learning Classifiers}\label{sec:bespoke}

The low-fabrication and non-recurring engineering (NRE) costs of printed circuits can be leveraged to build highly customized bespoke ML circuits, i.e., circuit is customized to a specific model trained on a specific training dataset.
Such degree of customization is not realizable in conventional silicon-based systems, due to their high NRE costs.
Following the design methodology of~\cite{Mubarik:MICRO:2020:printedml}, we examine four different ML classification algorithms (Fig.~\ref{fig:architectures}), i.e., Multi-Layer Perceptrons classifier (MLP-C), Multi-layer Perceptron regressor (MLP-R), Support Vector Machine classification (SVM-C), as well as Support Vector Machine regression (SVM-R), and evaluate them in terms of accuracy and hardware overheads in printed technologies.
In these customized models \yellow{all} coefficients are hardwired in the circuit's description/implementation
and thus, logic is further simplified by constant propagation etc.
The topology of MLP-C and MLP-R (Fig.\ref{fig:architectures}a,b) is composed of one hidden layer and one up to five neurons, so that, for each model, close to maximum accuracy is achieved with the least number of hidden nodes.
Moreover, ReLU activation functions are used.
SVMs (Fig.\ref{fig:architectures}c,d) use a linear kernel and SVM-Cs are implemented with 1-vs-1 classification.

Each algorithm is trained on six different datasets of the UCI ML repository~\cite{Dua:2019:uci} (see Table~\ref{tab:baselines})
These datasets are selected similar to~\cite{Mubarik:MICRO:2020:printedml,Weller:2021:printed_stoch} and could form representative examples of sensor-based printed applications.
Training is performed using scikit-learn and a hyperparameter search (RandomizedSearchCV) with 5-fold cross validation.
All input features are normalized to $[0,1]$ and each dataset is divided into a random $30\%/70\%$ test and training dataset respectively.
For our baseline bespoke circuits we consider fixed-point arithmetic with the precision of all inputs and coefficients set to 4-bit and 8-bit, respectively, i.e., the smallest precision in which accuracy is close to floating-point one for all models.
Synthesis and power estimation of all circuits is obtained from Synopsys tools using the open source Electrolyte Gated Transistor (EGT) library~\cite{Bleier:ISCA:2020:printedmicro}, while testing accuracy is acquired with circuit simulations using Questasim.
All the circuits are synthesized at a relaxed clock(i.e., 200ms for all the designs), targeting to further improve area efficiency.
Then, the obtained circuits are operated at the maximum sustainable frequency (minimum delay) in which no timing violations occur.

Table~\ref{tab:baselines} presents the characteristics, computation and hardware requirements (e.g. area and power) for our different baseline bespoke implementations.
As we can see, many circuits occupy area that is prohibitive for most printed applications ($>\!12cm^2$ on average), while power consumption of most circuits is so high (mainly $>30$mW) that they cannot be powered by a single existing printed battery.

\section{Cross-Layer Approximation for Printed ML Bespoke Circuits}
\orange{

\begin{figure}[t]
\centering
\includegraphics[]{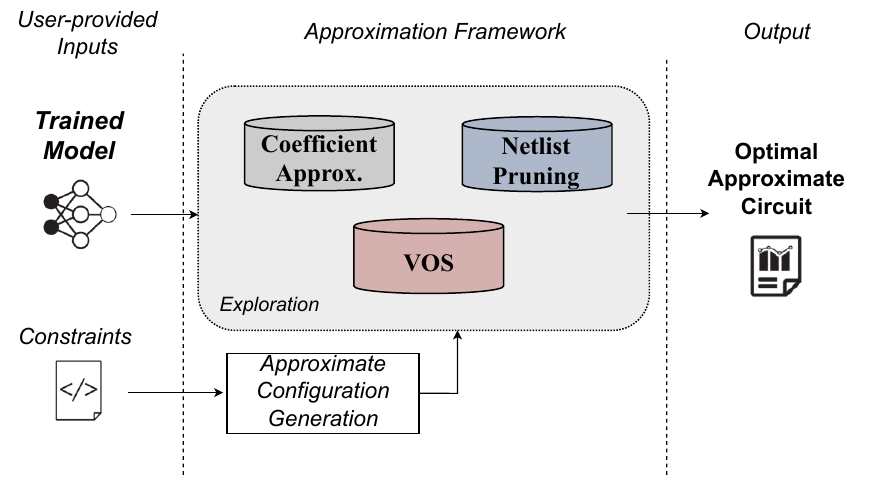}
\caption{Overview of our proposed framework flow diagram.
}
\label{fig:framework}
\end{figure}

In this section, we present our automated framework (Fig.~\ref{fig:framework}) for approximate printed ML circuits.
Briefly, our framework receives as input a trained model (e.g., dumped from scikit-learn) and performs a hardware-driven coefficient approximation.
Next, our framework prunes the generated synthesized netlist through a full search DSE by systematically prunning gates based on their significance and their switching activity.
Finally, to further boost power efficiency, a VOS exploration is performed in the derived Pareto space, where Pareto optimal approximate circuits are obtained.

\subsection{Hardware-Driven Coefficient Approximation}\label{sec:coa}

A weighted sum (as for example in the case of MLPs and SVMs) is expressed as:
\begin{equation}
    S=\sum_{1\leq i \leq N}{x_i\cdot w_i},
\end{equation}
where $w_i$ are the predefined coefficients (or weights) obtained after training, $x_i$ are the inputs, and $N$ is the number of coefficients.
In the case of bespoke ML architectures, these coefficients are hardwired within the circuit~\cite{Mubarik:MICRO:2020:printedml}.
As a result, the area (and power) of each bespoke multiplier $\mathrm{BM}_w$ required to compute the product $x\cdot w$, $\forall x$, is determined by the value of the coefficient $w$ and the width of the input $x$.
For example, Fig.~\ref{fig:mult8area} presents the area of $\mathrm{BM}_w$, $\forall w \in [-128,127]$ (i.e., 8-bit coefficients), for 4-bit and 8-bit input values.
For comparison, in the caption of Fig.~\ref{fig:mult8area} we also report the area of the conventional $4\times 8$ and $8 \times 8$ multipliers.
In both cases, the bespoke multipliers $\mathrm{BM}_w$ offer significantly lower area than the conventional multiplier for all the $w$ values.
Moreover, it is evident that the area of $\mathrm{BM}_w$ highly depends on $w$ and the input bitwidth.
However, similar trend is observed in Fig~\ref{fig:mult8area}a and~\ref{fig:mult8area}b, i.e., neighbouring $w$ values may offer significantly different area.
Importantly, in many cases the area may be nullified, e.g., when $w$ is a power of two.
This motivates us to investigate and propose a hardware-driven coefficient approximation, tailored for bespoke architectures, that replaces a coefficient value $w$ with a neighbouring value $\tilde{w}$ so that \texttt{AREA}($\mathrm{BM}_{\tilde{w}}$) $<$ \texttt{AREA}($\mathrm{BM}_w$).

Fig.~\ref{fig:multareasav} presents the area reduction that is achieved by our coefficient approximation with respect to several bespoke multipliers sizes (a-d).
To generate each boxplot in Fig.~\ref{fig:multareasav}, for all $w$, we select $\tilde{w}$ so that $\tilde{w}$ offers the lowest \texttt{AREA}($\mathrm{BM}_{\tilde{w}}$) and $\tilde{w} \in [w-e,w+e]$, where $e$ is a given threshold (x-axis).
Clipping is applied at the borders.
As shown in Fig.~\ref{fig:multareasav}, the obtained $\mathrm{BM}_{\tilde{w}}$ offer significantly lower area than the $\mathrm{BM}_w$.
Our coefficient approximation delivers a median area reduction of more than $19\%$ when $e=1$ while this value increases to $53\%$ when $e=4$.
Nevertheless, in most cases, for $e\geq4$ the area reduction becomes less significant.
For example, in Fig.~\ref{fig:multareasav}b, the median area reduction is $44\%$ for $e=4$ and increases to only $61\%$ for $e=10$.
In many cases, in Fig.~\ref{fig:multareasav}, the area reduction goes up to $100$\% or it is $0$\%.
The former is explained by the fact that $w$ was replaced by $\tilde{w}$ that was a power of two and thus the area reduction is $100$\% since $\tilde{w}$ features zero area.
On the other hand, in the cases that $w$ features the lowest area in the segment $[w-e,w+e]$, $w$ is not replaced and the area reduction is zero.

\begin{figure}[t!]
\centering
\includegraphics[]{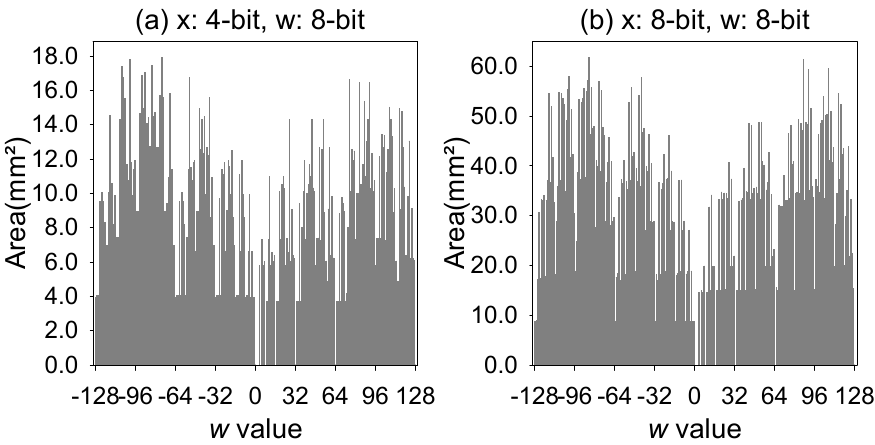}
\caption{The area of the bespoke multiplier w.r.t. the coefficient value $w$. Two architectures are considered: a) 4-bit inputs and 8-bit coefficients and b) 8-bit inputs and 8-bit coefficients.
For reference the area of the conventional $4\times 8$ and $8\times 8$ multipliers is 83.61$mm^{2}$ and 207.43$mm^{2}$, respectively. Figure obtained from~\cite{DATE22:Armen}.
}

\label{fig:mult8area}
\end{figure}

\begin{figure}[t]
\centering
\includegraphics[width=0.45\columnwidth]{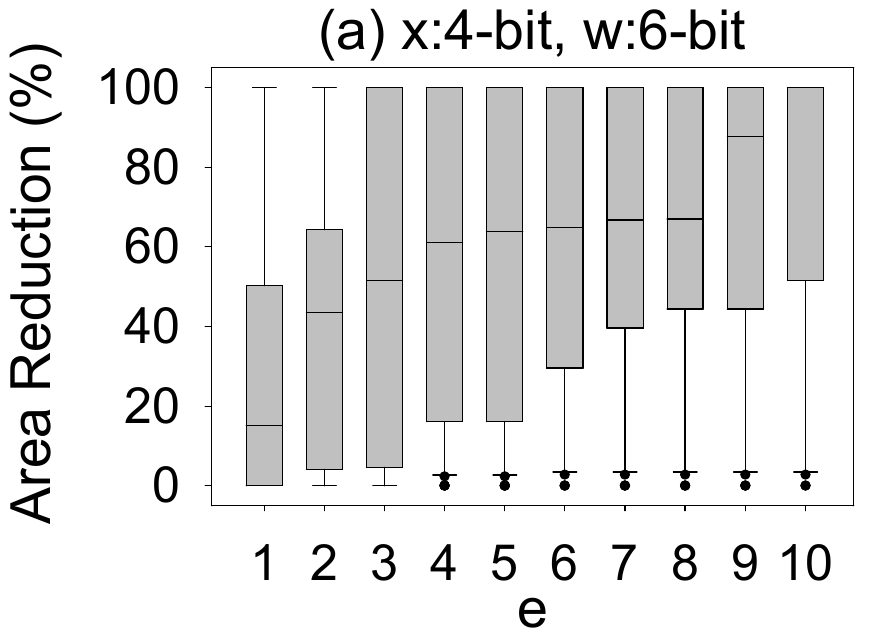}
\includegraphics[width=0.45\columnwidth]{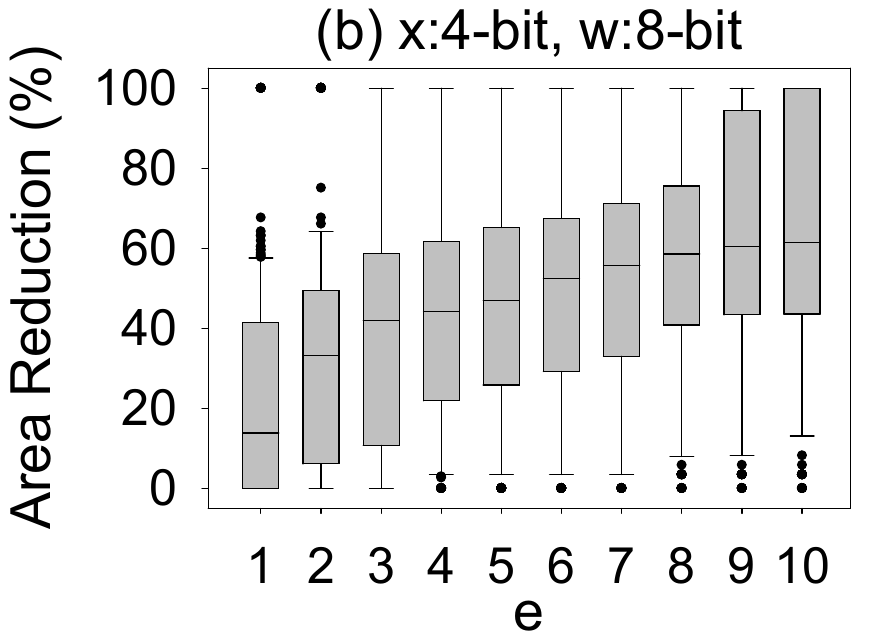}\\
\includegraphics[width=0.45\columnwidth]{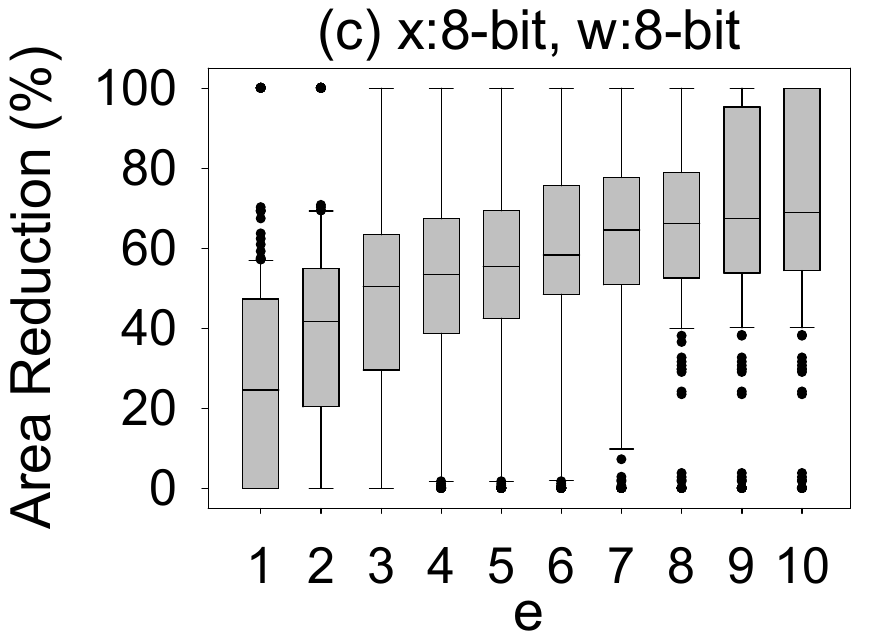}
\includegraphics[width=0.45\columnwidth]{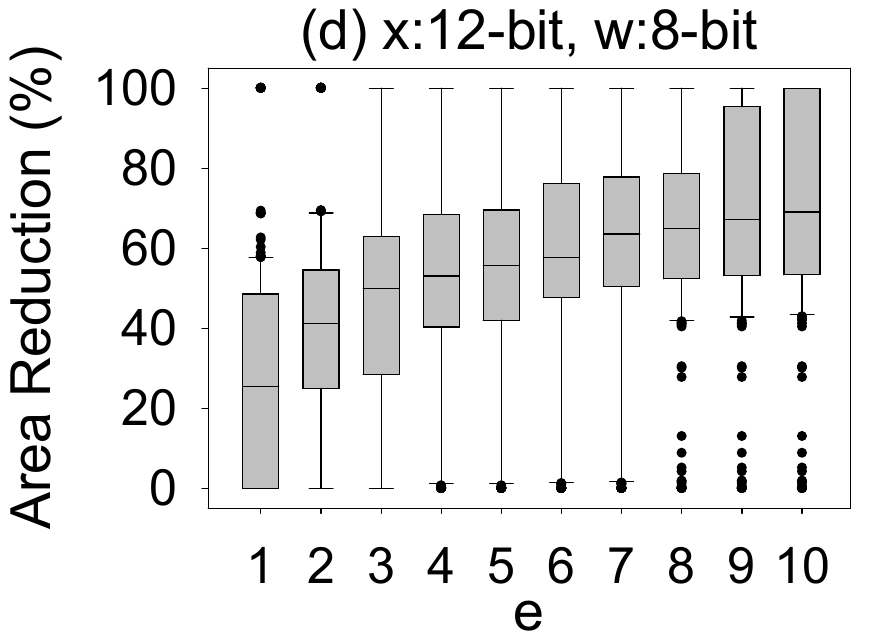}
\caption{The area reduction delivered by our coefficient approximation when $(w-\tilde{w})\leq e$. Several bespoke multipliers are considered (a-d). Figure obtained from~\cite{DATE22:Armen}.}
\label{fig:multareasav}
\end{figure}

When replacing $w$ by $\tilde{w}$, the multiplication error is equal to $x\cdot(w-\tilde{w})$.
Thus, the error $\epsilon_S$ of the weighted sum is:
\begin{equation}\label{eq:wsumerror}
    \epsilon_S =\sum_{1\leq i \leq N}{x_i\cdot (w_i-\tilde{w}_i). }
\end{equation}
Considering positive inputs (see Section \ref{sec:bespoke}), by systematically selecting $\tilde{w}_i$ to balance the positive and negative errors (i.e., $w_i-\tilde{w}_i$), we can minimize~\eqref{eq:wsumerror}.

Given an MLP or SVM, we implement our proposed hardware-driven coefficient approximation as follows:
\begin{enumerate}[leftmargin=*]
\item Given the coefficients $w_i$ and the bitwidth of the inputs, we evaluate the area of all the bespoke multipliers (\texttt{AREA}($\mathrm{BM}_{\tilde{w}}$)), $\forall i$ and $\forall \tilde{w} \in [w_i-e, w_i+e]$.
This step uses Synopsys Design Compiler and the EGT PDK~\cite{Bleier:ISCA:2020:printedmicro} for circuit synthesis and area analysis.~\label{item:synthbm}
\item For all the coefficients $w_i$, create a set $R_i=\{\tilde{w}_i^-,\tilde{w}_i^+\}$ s.t. $\tilde{w}_i^- \in [w, w+e]$ and  $\tilde{w}_i^-$ features the smallest area in that segment, i.e., \texttt{AREA}($\mathrm{BM}_{\tilde{w}_i^-}$) = min(\texttt{AREA}($\mathrm{BM}_{\tilde{w}_i}$)), $\forall \tilde{w} \in [w, w+e]$.
Similarly, we select $\tilde{w}_i^+ \in [w-e, w]$. By definition replacing $w$ with $\tilde{w}_i^-$ generates a negative error while replacing $w$ with $\tilde{w}_i^+$ generates a positive error.~\label{item:minarea}
\item We perform a brute-force search to select the configuration $\{\tilde{w}_i: \tilde{w}_i \in R_i, \forall i \}$ so that $\sum_{\forall i}{(w_i-\tilde{w}_i)}$ is minimized. In case of a tie, we select the one that minimizes $\sum_{\forall i}{\texttt{AREA}(\mathrm{BM}_{\tilde{w}_i})}$.
Note that given the small search space size, brute force approach is feasible.~\label{item:miner}
\end{enumerate}
Steps~\ref{item:synthbm}-\ref{item:miner} are executed for each weighted sum, i.e., neuron in MLPs and 1-vs-1 classifier in SVMs.
In addition, we set $e$=$4$ in our analysis since for $e>4$ the area gains quite saturate (see Fig.~\ref{fig:multareasav}).
In step~\ref{item:miner} we implement an exhaustive search to extract the final configuration.
Unlike conventional silicon VLSI, in printed electronics the examined ML models are rather small in size (in terms of number of parameters).
Hence, each weighted sum (neuron or classifier) features only a limited number of coefficients, i.e., the size of the design space is well constrained.
It is noteworthy, that in the worst case, step~\ref{item:miner} required only 3s using $80$ threads.
The aforementioned execution times refer to a dual-CPU Intel Xeon Gold 6138 server.

In our optimization (steps~\ref{item:synthbm}-\ref{item:miner}) the sum $\sum_{\forall i}{\texttt{AREA}(\mathrm{BM}_{\tilde{w}_i})}$ is used as a proxy of the area of the weighted sum.
In other words, by minimizing the area (through our coefficient approximation) of the required bespoke multipliers, we aim in minimizing the area of the weighted sum.
We evaluate our area proxy against $1000$ randomly generated weighted sum circuits (i.e., random coefficients and input sizes).
The Pearson correlation coefficient between the area of the weighted sum obtained by Design Compiler and the area estimation using $\sum_{\forall i}{\texttt{AREA}(\mathrm{BM}_{\tilde{w}_i})}$ is $0.91$, i.e., perfect linear correlation.
Hence, our proxy precisely captures the area trend and minimizing $\sum_{\forall i}{\texttt{AREA}(\mathrm{BM}_{\tilde{w}_i})}$ in our optimization, will result in a weighted sum circuit with minimal area.
Finally, since our technique replaces the coefficient values with approximate more hardware-friendly ones, it does not require any specific/custom hardware implementation (e.g., as usually done in logic approximation).
Hence, it can be seamlessly integrated in any design framework and exploits all the optimization and IPs (e.g., multipliers) of synthesis tools.

\subsection{Netlist Pruning}\label{subsec:prune}

\begin{figure}[t]
\centering
\includegraphics[]{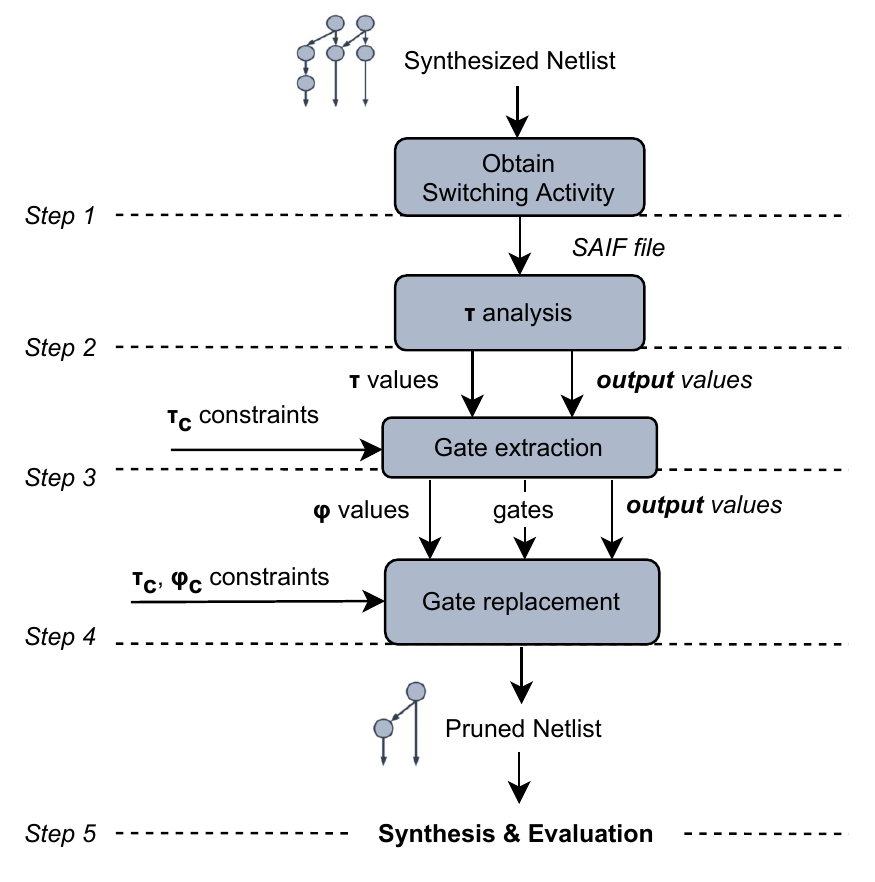}
\caption{Overview of our netlist pruning approach
}
\label{fig:netprun}
\end{figure}

To further increase the area efficiency, in addition to our coefficient approximation, we apply netlist pruning.
Netlist pruning is based on the observation that the output of several gates in a netlist remains constant (`0' or `1') for the majority of the execution time.
Hence, removing such a gate from the netlist and replacing its output with a constant value, results in low error rate.
Netlist pruning has been widely studied to enable design-agnostic approximation~\cite{GatePrun2017,Scarabottolo:DAC:2019:prune}. 
In this section we provide a brief description of how we implemented and tailored netlist pruning for bespoke printed ML architectures.

First we define two pruning parameters for a gate: $\tau$ is the maximum percentage of time that the gate's output is `0' or `1' and $\phi$ the most significant output bit (starting from 0) that the gate is connected to (through any path).
Using $\tau$ and $\phi$ \yellow{we constraint} the error frequency and the error magnitude, respectively. 
For example, assume that gate U1 is `1' the $\tau$=$90$\% of the time and that $\phi$=$3$.
Replacing U1 by `1' will result in an error rate of $10$\% and the maximum error will be less than $2^4$.
Netlist pruning is mainly implemented using heuristics~\cite{GatePrun2017} and thus optimality cannot be guaranteed.
In our work, leveraging that i) bespoke architectures feature significantly fewer area/gates than conventional architectures and ii) that ML models for printed electronics are rather limited in size, we use $\tau$ and $\phi$ to constraint the pruning design space and we implement an exhaustive search to obtain Pareto-optimal solutions.
Aiming for high area-efficiency, we prune all gates that feature $\tau$ and $\phi$ less or equal to given constraints $\tau_c$ and $\phi_c$.
Since all the pruned gates feature $\phi \leq \phi_c$, the maximum output error is less than $2^{\phi_c+1}$ irrespective of the number of pruned gates.
Overall, our coarse-grained approach ensures maximum area reduction while satisfying a maximum error threshold and enables fast design space exploration since only a gate's $\tau$ and $\phi$ need to be calculated.

Leveraging $\phi$ we filter all the gates that feature high $\tau$ and prune those that satisfy a given worst-case error.
In the case of regressors (MLP-R and SVM-R) this works well and many gates are pruned for low $\phi_c$ values.
However, classifiers require special consideration.
MLP-C and SVM-C use an argmax function at the end to translate the numerical predictions (e.g., values of output neurons in MLP-C) to a class.
As a result, the paths passing from all the gates are eventually congested in a few output bits, limiting the pruning granularity (possible $\phi_c$). 
Moreover, argmax breaks the correlation between the introduced numerical error in predictions and the final output.
For example, argmax([$0.9$, $0.1$])=argmax([$0.4$, $0.1$])=$0$.
For this reason, for the classifiers, we calculate $\phi$ for each gate with respect to the inputs of the argmax function.
For example, assume an MLP-C with $k$ output neurons $O_1$,..., $O_k$.
We define the value $\phi$ of a gate as $\max\limits_{\forall i}\phi(O_i)$, where $\phi(O_i)$ is the most significant output bit of the neuron $O_i$ that the gate is connected to.
If such a path doesn't exist, we set $\phi(O_i)=-1$.

Given a netlist (either exact or coefficient approximated), our netlist pruning operates as follows (also depicted in Fig.~\ref{fig:netprun}):
\begin{enumerate}[leftmargin=*]
\item Run RTL simulation of the synthesized netlist using the training dataset and Questasim to obtain the switching activity interchange format (SAIF) file.\label{item:sim}
\item Parse SAIF to calculate $\tau$ and the respective constant value (`0' or `1' ) for each gate.
For example, if the output of a gate is the $85$\% of the time `1' and the $15$\% it is `0', then $\tau$=$85$\% and the constant value is `1'. \label{item:tau}
\item Extract all the gates with $\tau \leq \tau_c$ and calculate their $\phi$. $\phi$ is easily calculated with the synthesis tool by reporting paths from a gate to the outputs.
\label{item:tauphi}
\item Prune all gates with $\tau \leq \tau_c$ and $\phi \leq \phi_c$, i.e., replace their output with the constant value extracted in step~\ref{item:tau}.\label{item:prune}
\item Synthesize the pruned netlist and evaluate its area and power as well as its accuracy on the test dataset.\label{item:synth}
\end{enumerate}
The pruned netlist is synthesized to exploit all optimizations of the synthesis tool, e.g., constant propagation.
Steps~\ref{item:sim} and~\ref{item:tau} are executed only once.
Step~\ref{item:tauphi} is executed $\forall \tau_c \in [80\%, 99\%]$.}
Then, for a given $\tau_c$, steps~\ref{item:prune}-\ref{item:synth} are executed $\forall \phi_c \in \Phi_\tau$.
$\Phi_\tau$ is equal to the set of the unique $\phi$ values obtained in step~\ref{item:tauphi}.
$\Phi_\tau$ enables us to explore only the relevant $\phi_c$ values.
\orange{
For example, if all the gates that feature $\tau\geq99$\% affect only the zero and first output bits, then $\phi_c>1$ is meaningless since it will return the same solution as $\phi_c=1$.
Unlike the pruning state-of-the-art that examines only very simple circuits~\cite{GatePrun2017,Scarabottolo:DAC:2019:prune}, our implementation is evaluated on complex ML circuits.
Moreover, as explained above for the classifiers, conventional pruning~\cite{GatePrun2017,Scarabottolo:DAC:2019:prune} cannot be used.}

\begin{figure}[t!]
\centering
\includegraphics[]{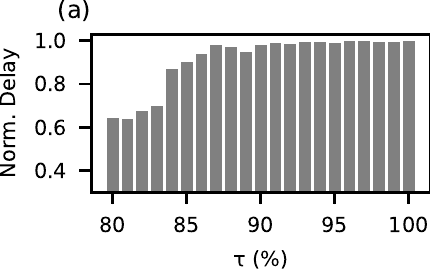}
\includegraphics[]{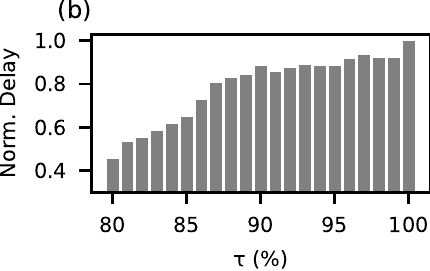}
\includegraphics[]{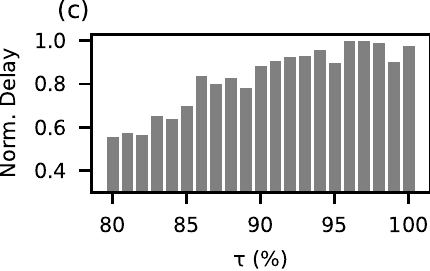}
\includegraphics[]{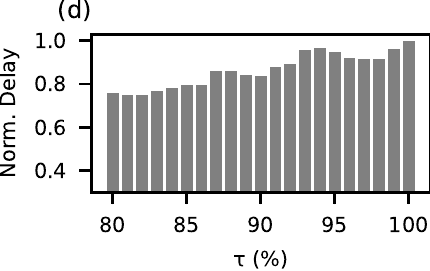}
\caption{Normalized critical path delay delivered by coefficient \& pruning approximation against exact bespoke design when $80\%\leq\tau\leq100\%$ for the same $\phi$ and for Seeds dataset. MLP-C (a), MLP-R (b), SVM-C (c) and SVM-R (d) are considered.
}
\label{fig:cpd}
\end{figure}

\subsection{Voltage Over-Scaling}\label{sec:vos}

Finally, at the circuit level we apply VOS to maximize the power gains of our framework since power consumption of a circuit depends quadratically on the voltage supply value:
\begin{equation}
    P_{total} = P_{static} + a \times C \times f_s \times V_{dd}^2,
\label{eq:vdd}
\end{equation} 
where $a$ is the switching activity, $C$ is the circuit's capacitance, $f_s$ is the operating clock frequency and $V_{dd}$ the voltage value.
It is evident, thus, that power consumption significantly decreases when voltage supply ($V_{dd}$) is decreased.

The main drawback of VOS is that circuits become slower and paths delay are increased. 
Therefore, when supply voltage is scaled below its nominal value~\cite{vader:zerv} for a given frequency, timing violations occur.
As the voltage continues to drop, an increasing number of internal paths cannot complete within the clock timing and the timing violation rate increases~\cite{vader:zerv}.

Applying coefficient approximation and netlist pruning not only reduces the circuit's area (and power) but also results in delay gain that can be therefore leveraged to apply aggressive VOS.   
An example of the obtained delay gain is illustrated in Fig.~\ref{fig:cpd}.
To generate Fig.~\ref{fig:cpd} we calculated for each model the delay when applying more aggressive approximation ($\tau<100\%$), normalized against the delay of their exact designs ($\tau=100\%$).
Note that $\phi$ was arbitrary selected equal to zero, since it does not affect circuit's logic, but only the error magnitude.
It is observed that, on average, $40\%$ lower delay can be achieved when $\tau=80\%$, while almost $9\%$ when $\tau$ is as high as $95\%$.
Hence, assuming iso-frequency operation for the approximate and baseline circuits, the approximate circuits will feature fewer timing violations (if any) and the impact of VOS on the delivered accuracy is expected to be significantly diminished.
As a result, considering that EGT printed circuits can be operated even down to $0.6V$~\cite{PrintVoltage:Tahoori},
combining VOS with coefficient approximation and netlist pruning, enables ultra-low power operation without any performance loss due to the voltage decrease.

In order to explore, in a timely manner\footnote{Significantly faster than SPICE simulations that would be infeasible to run as part of our optimization solution.}, the impact of VOS and evaluate the accuracy and power of voltage over-scaled approximate circuits, we perform VOS-aware gate-level post-synthesis timing simulations following the methodology presented in~\cite{vosim:zervakis}.
\yellow{Realistic voltage values of $1.0V$ down to $0.6V$~\cite{PrintVoltage:Tahoori} with a $20mV$ step are considered for the VOS application.
Each circuit is set to one optimized voltage, which is found using the presented offline analysis, and so no run-time voltage regulators or controllers are needed. 
Note also that printed batteries can be fully customized in terms of voltage, shape, polarity, etc.,~\cite{PrintedBatteries2018}.
}\label{commentR1C3b}
Finally, since VOS-errors are timing errors and thus input dependent~\cite{vader:zerv,vosim:zervakis}, large input datasets are required to efficiently capture the accuracy behavior of circuits under VOS.
For this reason, we create large input stimuli by replicating and shuffling each test dataset as many times as required in order to obtain 1M randomly sequenced inputs.
\yellow{In our evaluations we use the aforementioned simulation-based tool-flow built upon industrial-strength EDA tools (i.e., Synopsys Design Compiler, Prime Time, Mentor Questasim), using printed PDK and standard cell libraries~\cite{Bleier:ISCA:2020:printedmicro} calibrated based on fabrication and measurements from low voltage printing technologies.}\label{commentR1C2}



\subsection{Battery-driven Cross-Approximation}\label{sec:dse}

One of the main objectives of our work as well a key challenge in printed electronics is to enable battery-powered operation.
Hence, considering also that the space of available printed batteries is discrete and well constrained, without any loss of optimality, we assume a power constrained optimization for our cross-approximation.
In other words, the approximate printed ML classifier generated by our framework should be able to be powered by a specified printed battery. 
Given the area-accuracy efficiency of our coefficient approximation~\cite{DATE22:Armen}, we assume that all the approximate solutions generated by our framework employ coefficient approximation at the algorithmic level.
Then, at the logic and circuit levels we need to identify the respective approximation configuration, i.e., appropriate values for $\tau$ and $\phi$ (netlist pruning configuration) as well as $V_{dd}$ (VOS configuration), so that
i) the area is minimized,
ii) the accuracy is maximized, and 
iii) a power constraint is satisfied.
Naming $\axname$ the ML classifier after the coefficient approximation, our optimization problem can be formulated as follows:
\begin{equation}
\begin{gathered}
    \text{find} \, (\tau, \phi, V_{dd}) \, \text{s.t.} \\
    \mathrm{Power}\big(\axname(\tau, \phi, V_{dd})\big)\Big) \leq P_{BAT} \quad \text{and}  \\
    \min \Big(\mathrm{Area}\big(\axname(\tau, \phi, V_{dd})\big)\Big), \\
    \max \Big(\mathrm{Accuracy}\big(\axname(\tau, \phi, V_{dd})\big)\Big)
\end{gathered}
\label{eq:opt}
\end{equation} 
where is $P_{BAT}$ is the battery-specific power constraint. 
The size of the design space is from a few to several thousands of approximate circuits and the fact that each accuracy evaluation requires time consuming VOS-aware simulations exaggerates the complexity of our optimization problem. 
Note, however, that both the point-of-use fabrication process of printed circuits and the per model customization of bespoke architectures (i.e., our framework needs to run for every new ML circuit and/or targeted battery), mandate fast operation of our framework, in contrast to the long-lasting optimization cycles of silicon-based systems

To address this optimization problem, we first systematically reduce the design space and then employ a genetic algorithm to traverse the reduced design space.

\textbf{\textit{Pruning the design space:}}
Although assessing the accuracy of VOS-based approximate solutions is very slow, evaluating approximate circuits that apply only coefficient approximation and netlist pruning (i.e., $V_{dd}=1V$) is very fast (it requires a significantly smaller input stimuli and accuracy can be obtained through RTL simulations).\label{commentR1C1}
As a result, we implement a full search exploration evaluating the accuracy and power of all the approximate circuits $\axname(\tau, \phi, $1V$)$, $\forall \tau, \phi$.
First, we prune the design space by removing all the design points that feature accuracy loss higher than a certain percentage (e.g., $20\%$) as they are considered of poor quality.
Then, using~\eqref{eq:vdd} and the obtained $\mathrm{Power}(\axname(\tau, \phi,1V))$, we analytically estimate 
the $\mathrm{Power}(\axname(\tau, \phi, V_{dd}))$, $\forall \tau, \phi, V_{dd}$.
All the solutions that fail to satisfy the power constraint $ P_{BAT}$ are discarded and removed from the design space.

\textbf{\textit{Genetic Optimization:}} 
\yellow{To explore the pruned design space and find, quite fast, a close-to-optimal solution to our optimization problem, we employ a heuristic method~\cite{shafique} (Algorithm~\ref{alg:genetic}) based on NSGA-II~\cite{nsga2}.
Note that although genetic algorithms are proven to extract good enough solutions in complex optimization problems~\cite{shafique} any other heuristic algorithm can be seamlessly employed.}\label{commentR1C1a}
The approximation parameters are the \emph{genes} and, thus, the triplet $(\tau, \phi, V_{dd})$ represents the \emph{chromosome}.
The searching process starts by generating an initial population $\lambda$.
Instead of a totally random population, we guide the exploration by selecting configurations that satisfy specific accuracy and power constraints, based on some fast estimations.
This is done by firstly pre-calculating the candidate circuit's accuracy with the much faster RTL simulations (instead of the slow VOS-aware gate-level simulations) and then, by checking whether its power can satisfy the power constraint for any possible voltage value in the interval $[0.6V,\, 1.0V]$.
These estimations can quickly performed by using the theoretical power model of Eq.~\ref{eq:vdd}.
Therefore, ``non-acceptable'' configurations are removed and discarded from the design space.
After generating the initial population the following steps are repeated until the termination condition is satisfied: i) $\lambda$ offsprings are generated from initial population by means of usual mutation and crossover having similarly acceptable configurations, ii) $\lambda$ most fitting circuits (individuals) are selected, so that population remains at same initial size, iii) the viability of population known as \emph{fitness value} is evaluated.
Note that to find the best compromise between objectives that we are interested in (i.e., area, power and accuracy), fitness values are calculated using Kursawe function.
Finally, the process is terminated after $n$ epochs (generations of offsprings), where $n$ is an user defined constant tightly related to the model's characteristics and the constraints set by the user.

\begin{algorithm}[t!]
\caption{Pseudocode for proposed multi-objective optimization}\label{alg:genetic}
\footnotesize
\textbf{Input:} 1) Trained Model $m$, 2) Accuracy Loss Threshold $T$, \\3) Battery Power Threshold $P$, 4) Population\_Size $\lambda$, \\5) Termination Condition\\
\textbf{Output:} 1) Approx\_Configs, 2) Operating Voltage $v$\\
\begin{algorithmic}[1]
\State population $=$ POP\_INIT($m, \lambda$)
\State \textbf{while} Termination Condition \textbf{false}
\State \hspace{3mm} \textbf{do} 
\State \hspace{3mm} generate $\lambda$ offsprings
\State \hspace{3mm} population =+ offsprings
\State \hspace{3mm} \textbf{evaluate(}$\lambda$ population\textbf{)}
\State \textbf{return} Approx\_Configs , $v$
\vspace{3mm}
\State \textbf{function} POP\_INIT(model, pop\_size)
\State \hspace{3mm}pop\_n = 0
\State \hspace{3mm}\textbf{while} pop\_n $<$ pop\_size \textbf{do}
\State \hspace{6mm}Approx\_Configs = random\_init($\phi$, $\tau$, $v$)
\State \hspace{6mm}Approx Model $m^\prime$ = model(Approx\_Configs)
\State \hspace{6mm}$Acc_{m^\prime} \leftarrow $ RTL\_simulation
\State \hspace{6mm}\textbf{if} ($Acc_{m^\prime}$ $\geq$ $Acc_{m}$ $- T$)
\State \hspace{6mm}\hspace{3mm} $Pw_{m^\prime} \leftarrow$ power\_sim($1.00V$)
\State \hspace{6mm}\hspace{3mm} Estimate min $V_{dd}$ for given $P$ using \textbf{Eq.~\ref{eq:vdd}}
\State \hspace{6mm}\hspace{3mm} \textbf{if} $V_{dd} \geq 0.60V$ \textbf{then}
\State \hspace{6mm}\hspace{6mm} population =+ m(Approx\_Configs)
\State \hspace{6mm}\hspace{6mm} pop\_n =+ 1
\State \hspace{12mm} \textbf{end if}
\State \hspace{6mm}\textbf{end if}
\State \hspace{3mm}\textbf{endwhile}
\State \hspace{3mm}\textbf{return} population
\State \textbf{end function}
\end{algorithmic}
\end{algorithm}
\yellow{\section{Results \& Analysis}}\label{sec:results}\label{commentR1C2b}

\begin{figure*}[t!]
\centering
\includegraphics[]{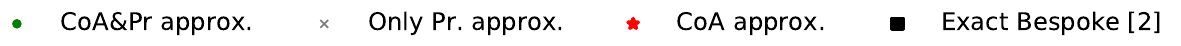}\\
\includegraphics[]{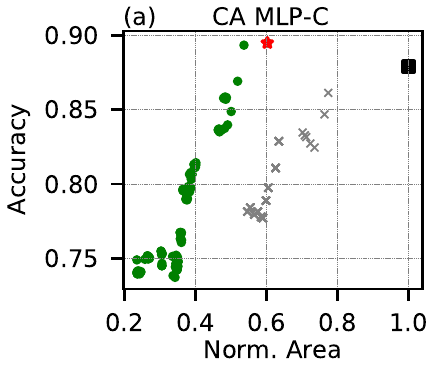}
\includegraphics[]{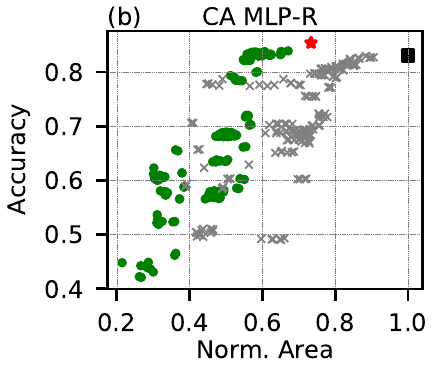}
\includegraphics[]{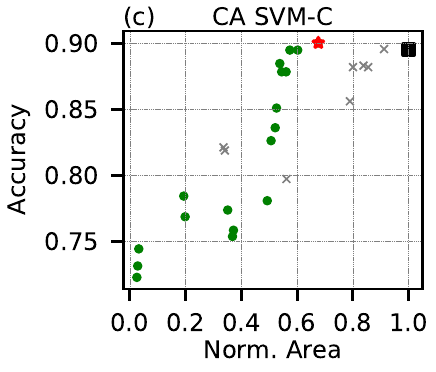}
\includegraphics[]{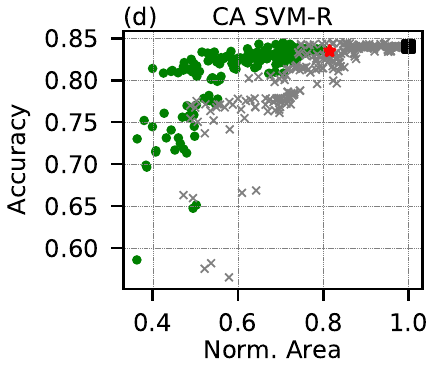}\\
\includegraphics[]{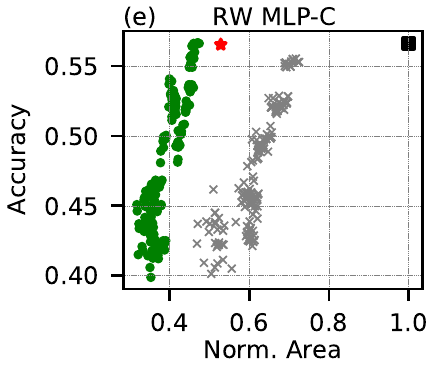}
\includegraphics[]{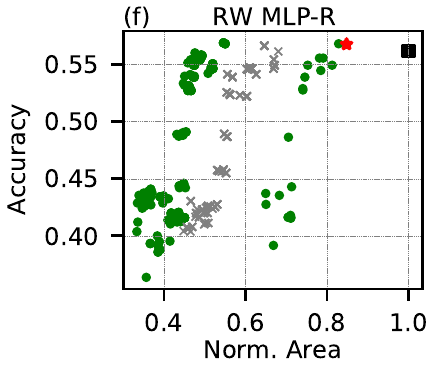}
\includegraphics[]{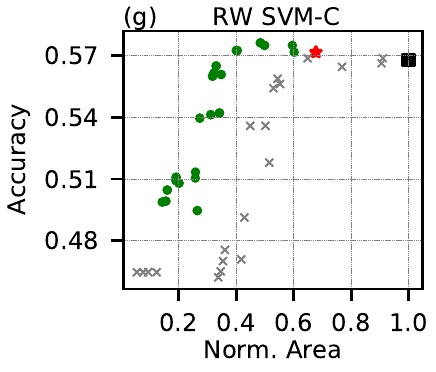}
\includegraphics[]{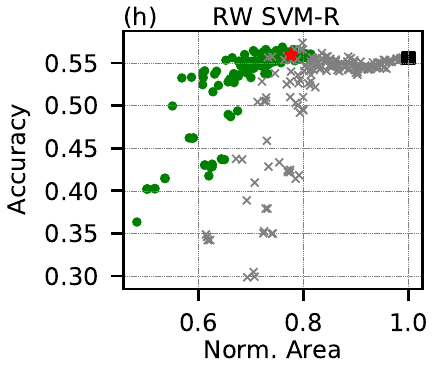}\\
\includegraphics[]{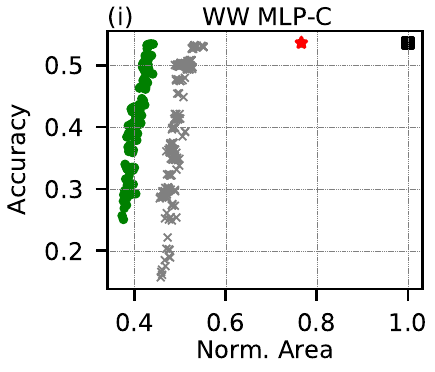}
\includegraphics[]{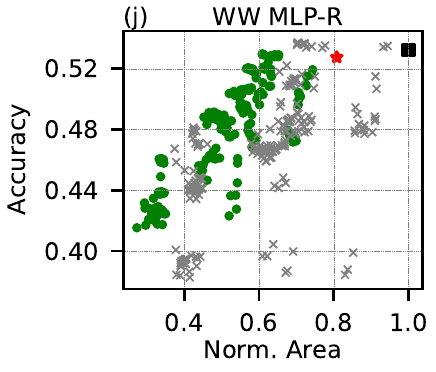}
\includegraphics[]{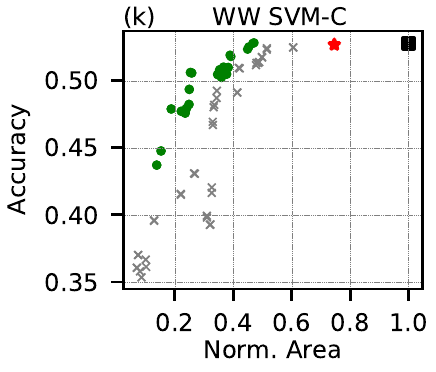}
\includegraphics[]{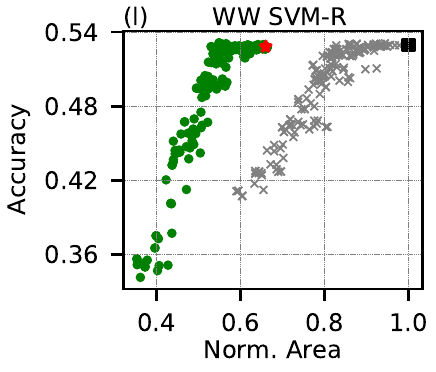}\\
\includegraphics[]{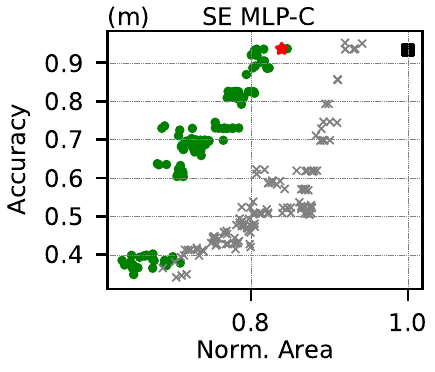}
\includegraphics[]{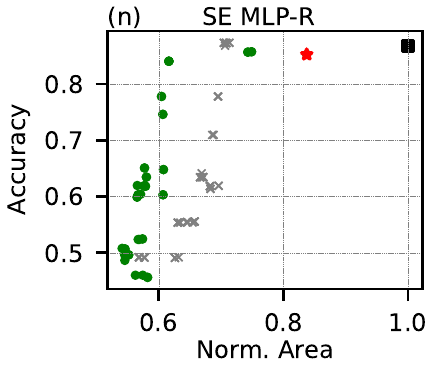}
\includegraphics[]{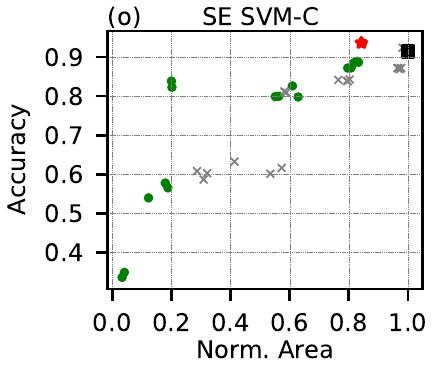}
\includegraphics[]{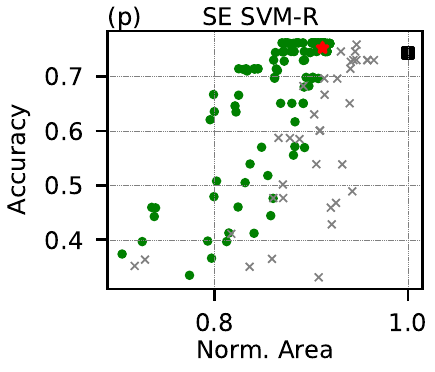}\\
\includegraphics[]{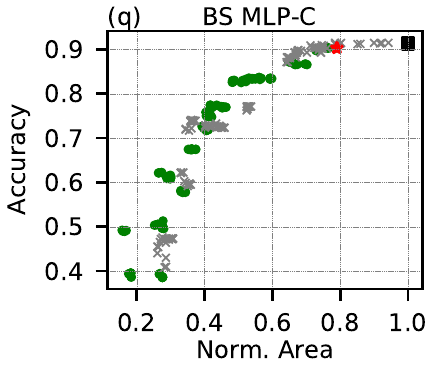}
\includegraphics[]{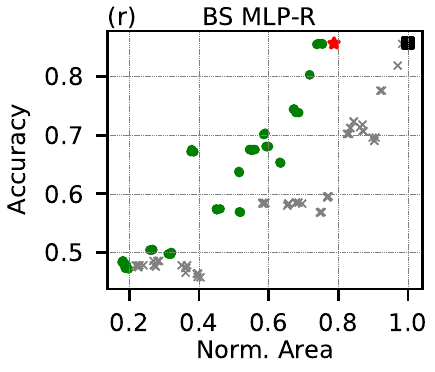}
\includegraphics[]{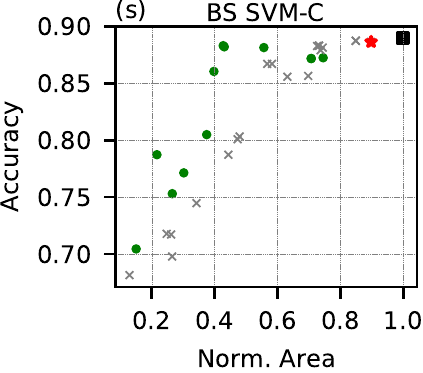}
\includegraphics[]{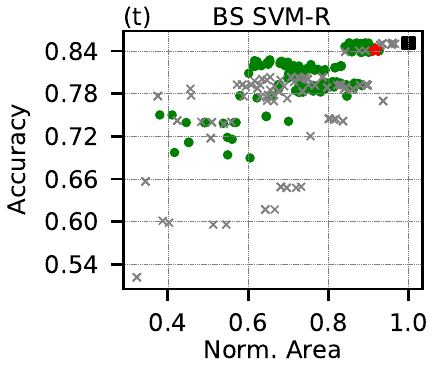}\\
\includegraphics[]{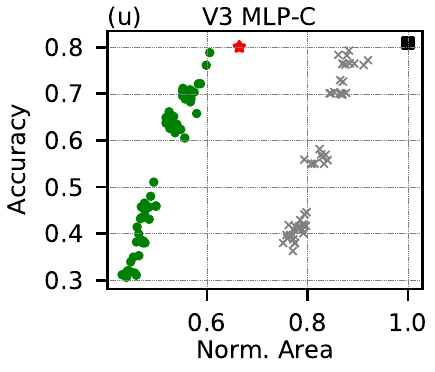}
\includegraphics[]{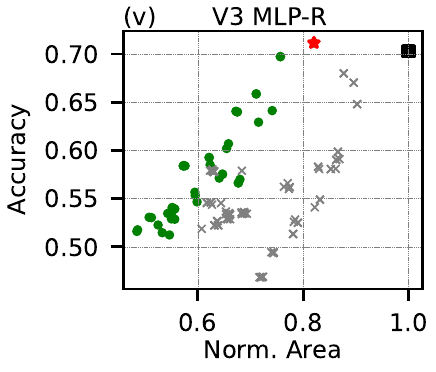}
\includegraphics[]{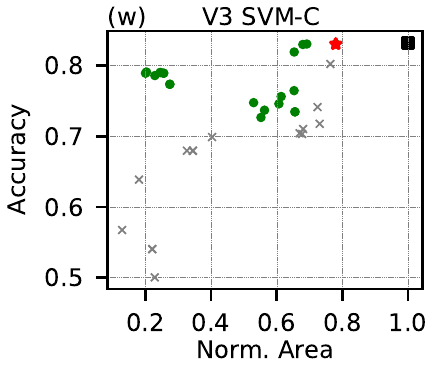}
\includegraphics[]{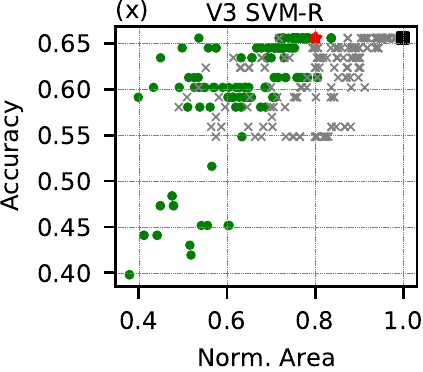}\\
\caption{Accuracy vs normalized area Pareto space. ML models examined: MLP-C, MLP-R, SVM-C, SVM-R for Cardio (a-d), RedWine (e-h), WhiteWine (i-l), Seeds (m-p), Balance Scale (q-t), Verterbal Column 3C (u-x), respectively.
}
\label{fig:dse}
\end{figure*}

\begin{figure*}[t!]
\centering
\includegraphics[]{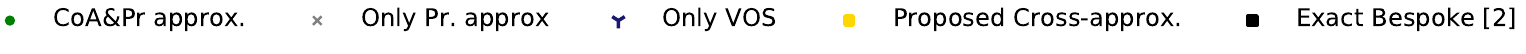}\\
\includegraphics[]{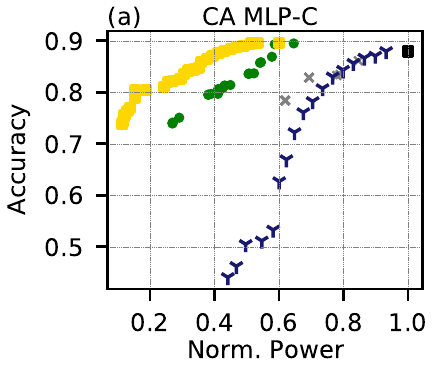}
\includegraphics[]{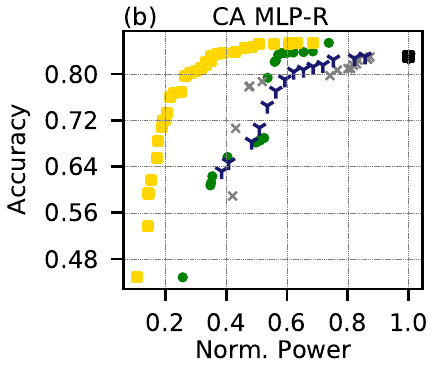}
\includegraphics[]{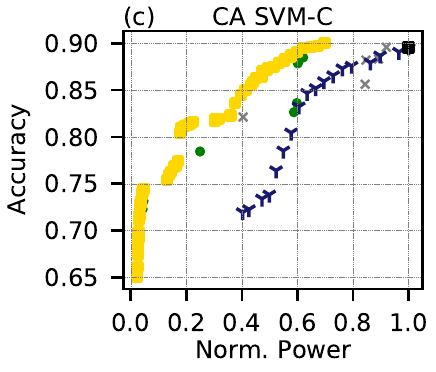}
\includegraphics[]{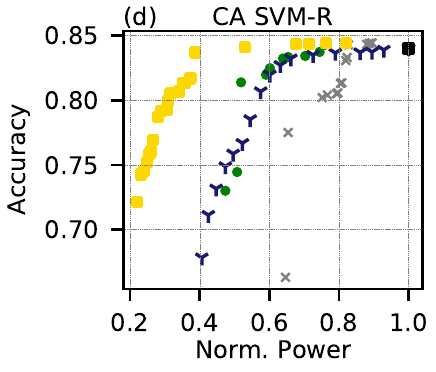}\\
\includegraphics[]{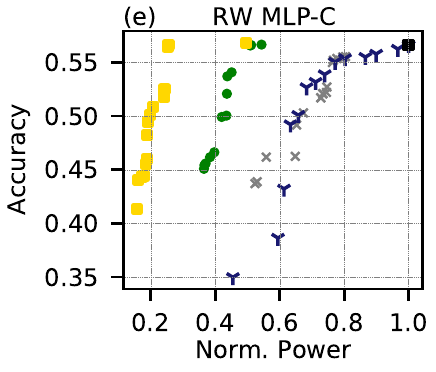}
\includegraphics[]{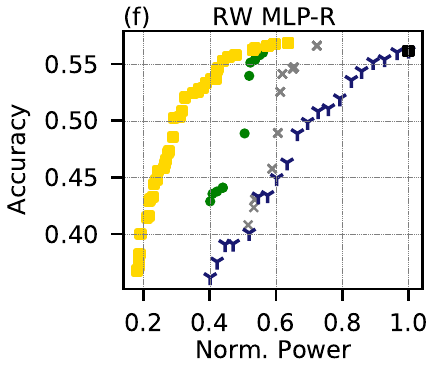}
\includegraphics[]{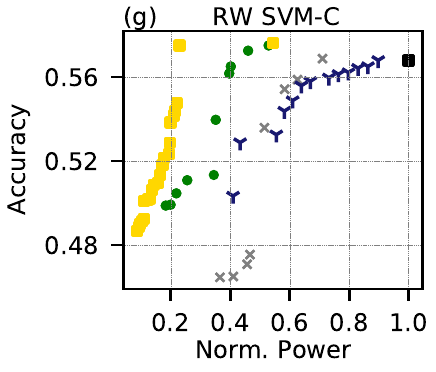}
\includegraphics[]{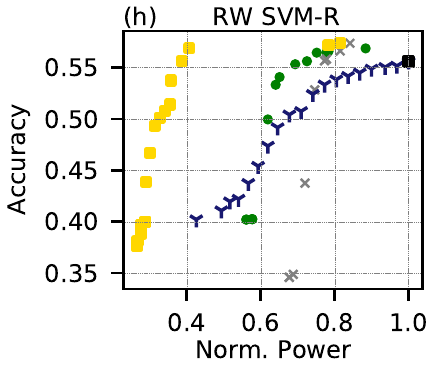}\\
\includegraphics[]{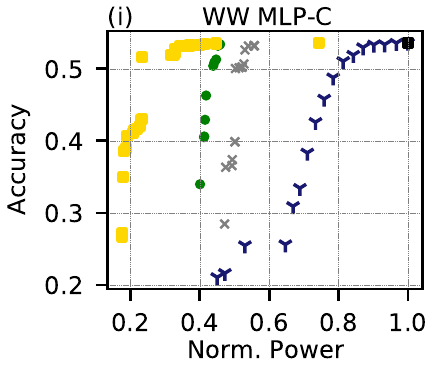}
\includegraphics[]{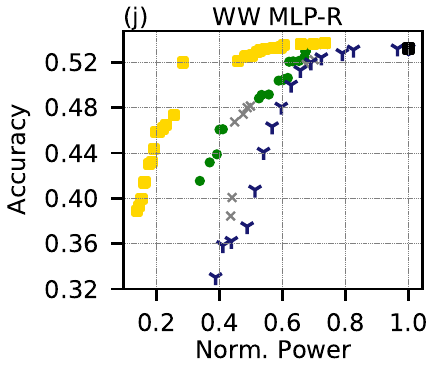}
\includegraphics[]{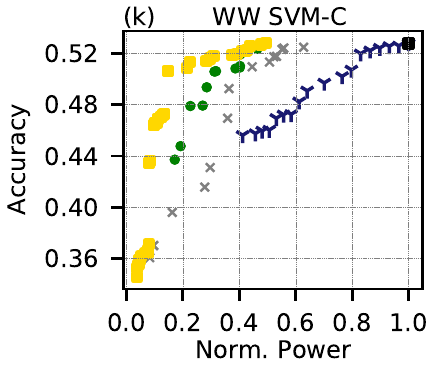}
\includegraphics[]{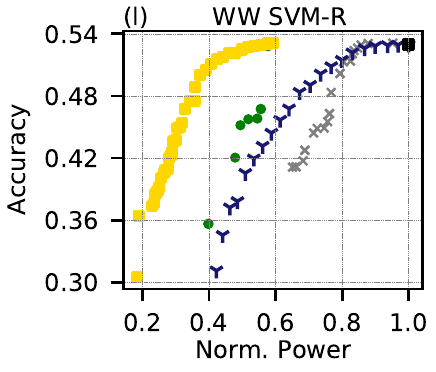}\\
\includegraphics[]{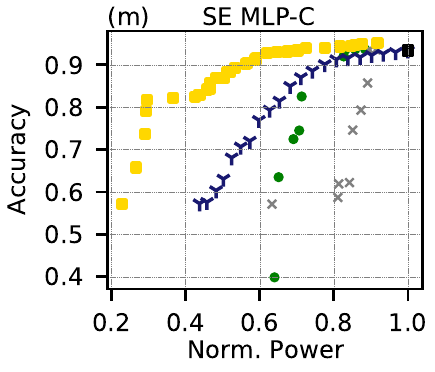}
\includegraphics[]{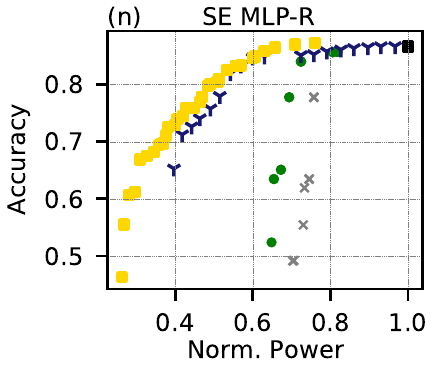}
\includegraphics[]{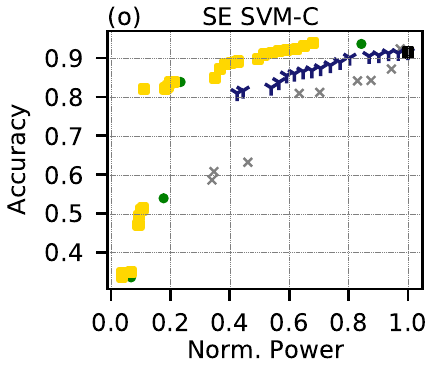}
\includegraphics[]{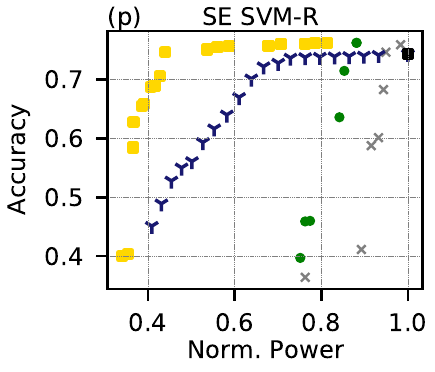}\\
\includegraphics[]{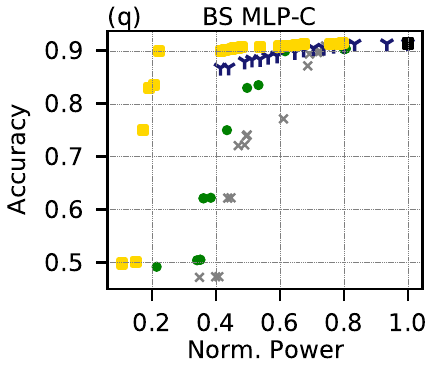}
\includegraphics[]{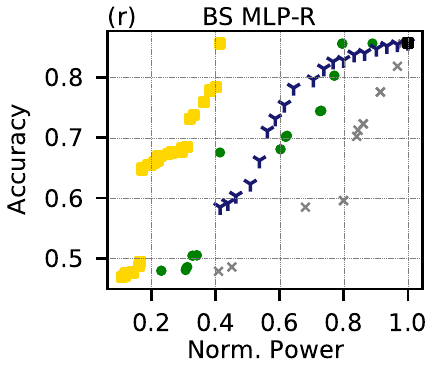}
\includegraphics[]{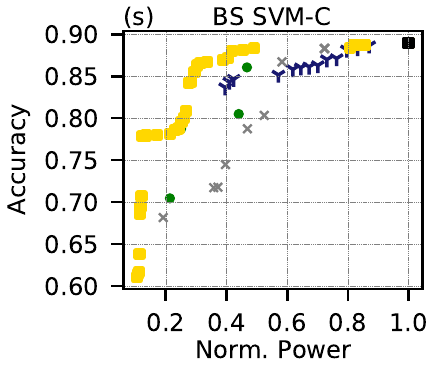}
\includegraphics[]{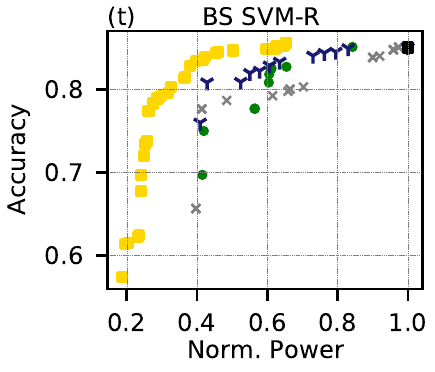}\\
\includegraphics[]{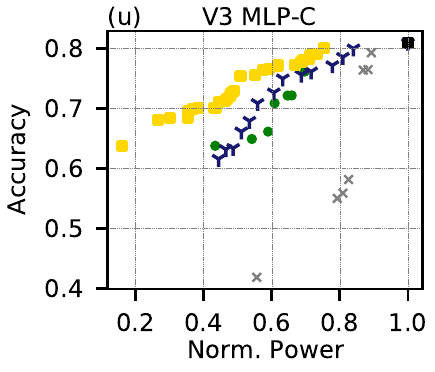}
\includegraphics[]{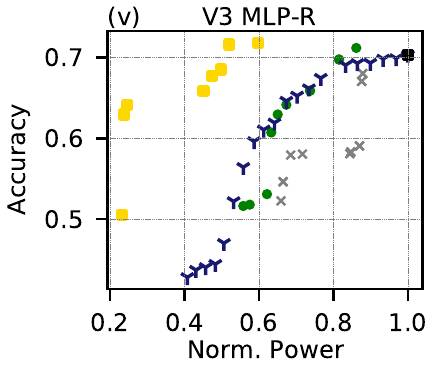}
\includegraphics[]{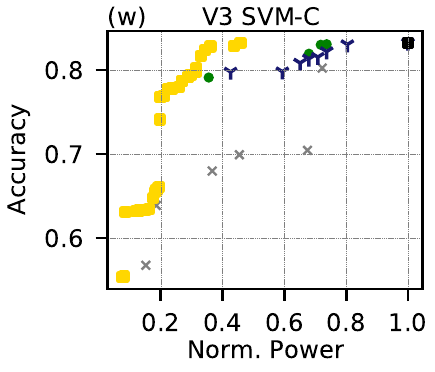}
\includegraphics[]{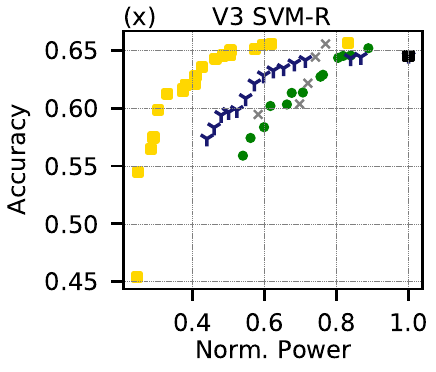}\\
\caption{Accuracy vs Normalized Power Pareto fronts for all evaluated techniques. ML models examined: MLP-C, MLP-R, SVM-C, SVM-R for Cardio (a-d), RedWine (e-h), WhiteWine (i-l), Seeds (m-p), Balance Scale (q-t), Vertebral Column 3C (u-x), respectively.
}
\label{fig:power}
\end{figure*}

This section an extensive evaluation of our work using the 24 ML classifiers presented in Table~\ref{tab:baselines}. 
First, we investigate and highlight the impact of approximate computing and specifically of our proposed cross-layer approximation in the design of printed ML classifiers.
Then, we evaluate our framework under varying battery constraints.

Fig.~\ref{fig:dse} presents the Accuracy-Area Pareto space for all the printed ML circuits examined (see Table~\ref{tab:baselines}).
In Fig.~\ref{fig:dse}, the black squares represent the baseline bespoke designs.
The red star is the design that applies only our proposed coefficient approximation.
The green dots are the designs that apply both Coefficient Approximation \& Pruning (``CoA\&Pr'').
In addition, Fig.~\ref{fig:dse} also presents in gray `x' the designs that use Only Pruning (``Only Pr'').
Only Pruning approximate circuits apply our pruning method directly on the baseline circuit and not on the coefficient approximated ones.
To obtain the Coefficient Approximation \& Pruning and the Only Pruning circuits, we performed a full-search design exploration.
Note that the number of the explored designs is circuit dependent.
As aforementioned, our pruning framework uses $80$\% $\leq \tau_c \leq$ $99$\% for all circuits but the explored $\phi_c$ values are circuit and $\tau_c$ dependent.
In total, to generate Fig.~\ref{fig:dse}, we evaluated more than $6000$ designs.

Overall, it is observed that approximate computing can effectively be employed to decrease the area complexity of the printed ML circuits since all the approximate designs feature lower area than the corresponding exact one (i.e., black square).
For example, for most circuits, more than $51$\% area reduction can be achieved for less than $5$\% accuracy loss.
\yellow{As shown in Fig.~\ref{fig:dse}, our coefficient approximation, by algorithmically balancing the positive and negative errors,
delivers\label{commentR2C2a} $22$\% average area reduction for almost identical accuracy with the exact baseline for all the examined circuits.}
In most cases it outperforms significantly the standalone gate-level pruning approximation.
In addition, designs generated using Coefficient Approximation \& Pruning (green dots) constitute mainly the area-optimal designs since they mainly form the Pareto front in all the sub-figures of Fig.~\ref{fig:dse}.
\yellow{It is also noteworthy that overall regressors are  more resilient to pruning than the respective classifiers.
The latter is explained by the $\phi$ parameter.
In the case of regressors a small $\phi$ results in some error in the least significant decimal digits that mainly have minimal impact on the selected class.
On the other hand, in classifiers even a smaller $\phi$ may break an inequality and, thus, directly lead to erroneous classification.
}\label{commentR2C2}

\begin{table*}[t!]
\caption{Area and power evaluation of all presented techniques for less than 1\% accuracy loss.}
\label{tab:threshold}
\setlength\tabcolsep{5.4pt}
\renewcommand{\arraystretch}{1.14}
\begin{threeparttable}
\begin{tabular}{l|ccccccccc?l|lllllllcl}
\thickhline
\multirow{2}{*}{\textbf{Model}} & \multicolumn{2}{c}{\textbf{Coa\&Pr\tnote{1}}} & \multicolumn{2}{c}{\textbf{CoA}} & \multicolumn{2}{c}{\textbf{Only Pr\tnote{2}}} & \multicolumn{3}{c?}{\textbf{Cross}} & \multirow{2}{*}{\textbf{Model}} & \multicolumn{2}{c}{\textbf{Coa\&Pr\tnote{1}}} & \multicolumn{2}{c}{\textbf{CoA}} & \multicolumn{2}{c}{\textbf{Only Pr\tnote{2}}} & \multicolumn{3}{c}{\textbf{Cross}} \\ \cline{2-10} \cline{12-20} 
 & \textbf{AG\tnote{3}} & \textbf{PG\tnote{3}} & \textbf{AG\tnote{3}} & \textbf{PG\tnote{3}} & \textbf{AG\tnote{3}} & \textbf{PG\tnote{3}} & \textbf{AG\tnote{3}} & \textbf{PG\tnote{3}} & \textbf{V{dd}} &  & \multicolumn{1}{c}{\textbf{AG\tnote{3}}} & \multicolumn{1}{c}{\textbf{PG\tnote{3}}} & \multicolumn{1}{c}{\textbf{AG\tnote{3}}} & \multicolumn{1}{c}{\textbf{PG\tnote{3}}} & \multicolumn{1}{c}{\textbf{AG\tnote{3}}} & \multicolumn{1}{c}{\textbf{PG\tnote{3}}} & \multicolumn{1}{c}{\textbf{AG\tnote{3}}} & \textbf{PG\tnote{3}} & \multicolumn{1}{c}{\textbf{Vdd}} \\ \cline{1-10} \cline{11-20} 
\textbf{CA MLP-C} & 48 & 44 & 40 & 36 & 0 & 0 & 48 & 61 & 0.74 & \textbf{SE MLP-C} & 16 & 14 & 10 & 9 & 8 & 10 & 16 & 38 & 0.76 \\
\textbf{CA MLP-R} & 45 & 44 & 27 & 26 & 16 & 15 & 45 & 65 & 0.88 & \textbf{SE MLP-R} & 29 & 25 & 16 & 14 & 26 & 23 & 29 & 37 & 0.60 \\
\textbf{CA SVM-C} & 43 & 38 & 33 & 29 & 9 & 8 & 43 & 43 & 0.60 & \textbf{SE SVM-C} & 16 & 16 & 15 & 16 & 2 & 2 & 16 & 48 & 0.88 \\
\textbf{CA SVM-R} & 49 & 42 & 19 & 15 & 26 & 22 & 49 & 55 & 0.60 & \textbf{SE SVM-R} & 12 & 11 & 9 & 6 & 7 & 5 & 12 & 56 & 0.60 \\
\textbf{RW MLP-C} & 55 & 50 & 47 & 43 & 0 & 0 & 55 & 75 & 0.82 & \textbf{BS MLP-C} & 24 & 20 & 21 & 18 & 25 & 21 & 24 & 55 & 0.60 \\
\textbf{RW MLP-R} & 53 & 49 & 15 & 14 & 35 & 30 & 53 & 58 & 0.72 & \textbf{BS MLP-R} & 26 & 21 & 21 & 17 & 2 & 2 & 26 & 59 & 0.82 \\
\textbf{RW SVM-C} & 68 & 65 & 32 & 31 & 35 & 33 & 68 & 77 & 0.80 & \textbf{BS SVM-C} & 52 & 44 & 11 & 13 & 19 & 17 & 52 & 52 & 0.90 \\
\textbf{RW SVM-R} & 35 & 33 & 22 & 22 & 27 & 25 & 35 & 61 & 0.76 & \textbf{BS SVM-R} & 15 & 15 & 8 & 5 & 7 & 4 & 15 & 55 & 0.60 \\
\textbf{WW MLP-C} & 57 & 57 & 23 & 26 & 47 & 48 & 57 & 67 & 0.60 & \textbf{V3 MLP-C} & 34 & 25 & 34 & 25 & 0 & 0 & 34 & 25 & 0.60 \\
\textbf{WW MLP-R} & 39 & 35 & 20 & 17 & 30 & 28 & 39 & 52 & 0.90 & \textbf{V3 MLP-R} & 24 & 19 & 18 & 14 & 0 & 0 & 24 & 48 & 0.64 \\
\textbf{WW SVM-C} & 61 & 59 & 26 & 25 & 49 & 47 & 61 & 69 & 1.00 & \textbf{V3 SVM-C} & 32 & 28 & 22 & 22 & 0 & 0 & 32 & 66 & 0.76 \\
\textbf{WW SVM-R} & 47 & 45 & 34 & 32 & 19 & 19 & 47 & 54 & 0.64 & \textbf{V3 SVM-R} & 20 & 18 & 4 & 4 & 26 & 24 & 20 & 57 & 0.98\\
\thickhline
\end{tabular}
\begin{tablenotes}\footnotesize
\item[] $^1$Proposed technique of our preliminary version~\cite{DATE22:Armen}. $^2$Proposed gate-level pruning technique. $^3$Area and Power gains compared to bespoke baseline~\cite{Mubarik:MICRO:2020:printedml} (in \%).
 
\end{tablenotes}
\end{threeparttable}
\end{table*}

Fig.~\ref{fig:power} assesses the Accuracy-Power trade-off of the approximate printed ML classifiers.
As a result, VOS is considered in Fig.~\ref{fig:power}.
Green dots and gray ``x'' represent now only the respective Pareto-optimal designs, while the golden squares form the Pareto-front design when considering our holistic model-to-circuit cross-approximation that comprises coefficient approximation, netlist pruning, and VOS (Ours).
Finally, blue ``tri down'' points are the voltage over-scaled circuits derived from the exact design.
Each Pareto front is obtained through an exhaustive design space exploration, i.e., we evaluated the accuracy and power consumption of all the circuits of Fig.~\ref{fig:dse} for all the voltage values considered (i.e., $\forall V_{dd} \in [0.6V, 1.0V]$, with a $20mV$ step). As shown in Fig.~\ref{fig:power}, the power savings of the approximate circuits that don't apply VOS (gray ``x''), follow a similar trend as the area gains in Fig.~\ref{fig:dse}.
As demonstrated by the full-search exploration of Fig.~\ref{fig:power}, our cross-approximation delivers the optimal results in terms of accuracy-power trade-off. 
Our approximate circuits feature $56\%$ power reduction for almost identical accuracy ($<0.01$) with the baseline circuits.
\yellow{Note also that VOS mainly affects the MSBs and, thus, when VOS errors occur they are high in magnitude~\cite{vosim:zervakis}, leading to poor prediction for regressors. 
On the other hand, in classifiers correct classification may be still obtained due to the argmax function that breaks the relation between numerical and classification accuracy and may potentially absorb the impact of such large errors.}\label{commentR2C2b}

\begin{table}[t!]
\centering
\caption{Area and accuracy evaluation of our genetic algorithm with power and accuracy constraints enabled. Three printed batteries are examined with 15\% accuracy loss threshold. Highlighted green designs have $\leq1\%$ accuracy loss.}
\label{tab:batteries}
\renewcommand{\arraystretch}{1.1}
\setlength\tabcolsep{4.5pt}
\begin{threeparttable}
\begin{tabular}{lcccccccc}
\hline \hline
\multicolumn{1}{l|}{\textbf{}} & \multicolumn{2}{c|}{\textbf{MLP-C}} & \multicolumn{2}{c|}{\textbf{MLP-R}} & \multicolumn{2}{c|}{\textbf{SVM-C}} & \multicolumn{2}{c}{\textbf{SVM-R}} \\ \cline{2-9} 
\multicolumn{1}{l|}{\textbf{}} & \textbf{\begin{tabular}[c]{@{}c@{}}Acc\\ Loss\\ (\%)\end{tabular}} & \multicolumn{1}{c|}{\textbf{\begin{tabular}[c]{@{}c@{}}Area\\ Gains\\ (\%)\end{tabular}}} & \textbf{\begin{tabular}[c]{@{}c@{}}Acc\\ Loss\\ (\%)\end{tabular}} & \multicolumn{1}{c|}{\textbf{\begin{tabular}[c]{@{}c@{}}Area\\ Gains\\ (\%)\end{tabular}}} & \textbf{\begin{tabular}[c]{@{}c@{}}Acc\\ Loss\\ (\%)\end{tabular}} & \multicolumn{1}{c|}{\textbf{\begin{tabular}[c]{@{}c@{}}Area\\ Gains\\ (\%)\end{tabular}}} & \textbf{\begin{tabular}[c]{@{}c@{}}Acc\\ Loss\\ (\%)\end{tabular}} & \textbf{\begin{tabular}[c]{@{}c@{}}Area\\ Gains\\ (\%)\end{tabular}} \\ \hline
{\textbf{}} & \multicolumn{8}{c}{\textbf{Molex 30mW \batd}} \\ \hline 
\multicolumn{1}{l|}{\textbf{CA}} & 7.5 & \multicolumn{1}{c|}{54} & \hlc1.0 & \multicolumn{1}{c|}{\hlc45} & 2.5 & \multicolumn{1}{c|}{46} & \hlc1.0 & \hlc49 \\
\multicolumn{1}{l|}{\textbf{RW}} & \hlc0.6 & \multicolumn{1}{c|}{\hlc55} & \hlc0.2 & \multicolumn{1}{c|}{\hlc53} & \hlc0.5 & \multicolumn{1}{c|}{\hlc68} & \hlc0 & \hlc20 \\
\multicolumn{1}{l|}{\textbf{WW}} & 11.9 & \multicolumn{1}{c|}{60} & \hlc0 & \multicolumn{1}{c|}{\hlc30} & 1.7 & \multicolumn{1}{c|}{61} & \hlc0 & \hlc45 \\
\multicolumn{1}{l|}{\textbf{SE}} & \hlc0.9 & \multicolumn{1}{c|}{\hlc16} & \hlc0 & \multicolumn{1}{c|}{\hlc28} & \hlc0 & \multicolumn{1}{c|}{\hlc16} & \hlc0 & \hlc9 \\
\multicolumn{1}{l|}{\textbf{V3}} & 4.1 & \multicolumn{1}{c|}{39} & \hlc0.2 & \multicolumn{1}{c|}{\hlc18} & \hlc0.8 & \multicolumn{1}{c|}{\hlc21} & \hlc0.3 & \hlc11 \\
\multicolumn{1}{l|}{\textbf{BS}} & \hlc0 & \multicolumn{1}{c|}{\hlc20} & \hlc0.6 & \multicolumn{1}{c|}{\hlc20} & \hlc0 & \multicolumn{1}{c|}{\hlc21} & \hlc0 & \hlc14 \\ 
\hline
{\textbf{}} & \multicolumn{8}{c}{\textbf{Zinergy 15mW \batb}} \\ \hline
\multicolumn{1}{l|}{\textbf{CA}} & 12.8 & \multicolumn{1}{c|}{73} & 12.2 & \multicolumn{1}{c|}{55} & 8.4 & \multicolumn{1}{c|}{66} & \hlc0 & \hlc26 \\
\multicolumn{1}{l|}{\textbf{RW}} & 5.9 & \multicolumn{1}{c|}{60} & \hlc0.2 & \multicolumn{1}{c|}{\hlc53} & 6.0 & \multicolumn{1}{c|}{80} & \hlc0 & \hlc20 \\
\multicolumn{1}{l|}{\textbf{WW}} &{-\tnote{1}}& \multicolumn{1}{c|}{-\tnote{1}} & \hlc1.0 & \multicolumn{1}{c|}{\hlc37} & 5.8 & \multicolumn{1}{c|}{81} & \hlc0 & \hlc45 \\
\multicolumn{1}{l|}{\textbf{SE}} & 12.0 & \multicolumn{1}{c|}{22} & 11.0 & \multicolumn{1}{c|}{28} & 2.9 & \multicolumn{1}{c|}{16} & \hlc0 & \hlc9 \\
\multicolumn{1}{l|}{\textbf{V3}} & 13.4 & \multicolumn{1}{c|}{44} & 7.9 & \multicolumn{1}{c|}{26} & \hlc0.8 & \multicolumn{1}{c|}{\hlc21} & \hlc0.3 & \hlc11 \\
\multicolumn{1}{l|}{\textbf{BS}} & 1.1 & \multicolumn{1}{c|}{27} & \hlc0.4 & \multicolumn{1}{c|}{\hlc21} & \hlc0 & \multicolumn{1}{c|}{\hlc21} & \hlc0 & \hlc14 \\
\hline
\textbf{} & \multicolumn{8}{c}{\textbf{Blue Spark 6mW \bata}} \\ \hline 
\multicolumn{1}{l|}{\textbf{CA}} &{-\tnote{1}}& \multicolumn{1}{c|}{-\tnote{1}} &{-\tnote{1}}& \multicolumn{1}{c|}{-\tnote{1}} &{-\tnote{1}}& \multicolumn{1}{c|}{-\tnote{1}} &{-\tnote{1}}&{-\tnote{1}}\\
\multicolumn{1}{l|}{\textbf{RW}} &{-\tnote{1}}& \multicolumn{1}{c|}{-\tnote{1}} &{-\tnote{1}}& \multicolumn{1}{c|}{-\tnote{1}} &{-\tnote{1}}& \multicolumn{1}{c|}{-\tnote{1}} & \hlc0 & \hlc20 \\
\multicolumn{1}{l|}{\textbf{WW}} &{-\tnote{1}}& \multicolumn{1}{c|}{-\tnote{1}} &{-\tnote{1}}& \multicolumn{1}{c|}{-\tnote{1}} &{-\tnote{1}}& \multicolumn{1}{c|}{-\tnote{1}} & 13.0 & 56 \\
\multicolumn{1}{l|}{\textbf{SE}} &{-\tnote{1}}& \multicolumn{1}{c|}{-\tnote{1}} &{-\tnote{1}}& \multicolumn{1}{c|}{-\tnote{1}} & 9.9 & \multicolumn{1}{c|}{80} &{-\tnote{1}}&{-\tnote{1}}\\
\multicolumn{1}{l|}{\textbf{V3}} &{-\tnote{1}}& \multicolumn{1}{c|}{-\tnote{1}} &{-\tnote{1}}& \multicolumn{1}{c|}{-\tnote{1}} & 6.2 & \multicolumn{1}{c|}{74} & 3.2 & 23 \\
\multicolumn{1}{l|}{\textbf{BS}} &{-\tnote{1}}& \multicolumn{1}{c|}{-\tnote{1}} &{-\tnote{1}}& \multicolumn{1}{c|}{-\tnote{1}} & \hlc0.7 & \multicolumn{1}{c|}{\hlc57} & \hlc0 & \hlc14 \\
\hline \hline
\end{tabular}
\begin{tablenotes}\footnotesize
\item[] $^1$ Cannot be powered by specific printed battery.
\end{tablenotes}
\end{threeparttable}
\end{table}

Table~\ref{tab:threshold}, considers a conservative accuracy loss threshold (i.e., only $1$\%) and 
reports the area and power gains of the area-power-optimal circuits for each approximation technique.
As shown, compared to the baseline bespoke~\cite{Mubarik:MICRO:2020:printedml} implementations, Coa\&Pr delivers on average $38\%$ and $34\%$ area and power reduction, respectively.
The corresponding values of CoA only are $22\%$ and $20\%$.
Ony pruning approximation achieves only $17\%$ and $16\%$ average area and power reduction, respectively.
On the other hand, our framework features $38\%$ area and $56\%$ power reduction.
It is noteworthy that the delay gain due to logic and algorithmic approximation (coefficient approximation and netlist pruning) enables operating the approximate circuits down to $0.74V$ (on average) while still satisfying tight accuracy loss constraints (i.e., $1$\%).
Hence, compared to Coa\&Pr~\cite{DATE22:Armen}, our cross approximation delivers an additional $22\%$ power reduction and same area savings. 

Next, we evaluate our framework in generating battery-constrained approximate printed ML classifiers of high accuracy.
For this evaluation we consider three different printed batteries: Molex 30mW, Zinergy 15mW, and Blue Spark 6mW.
In addition, we search for solutions with up to 15\% accuracy loss.
Table~\ref{tab:batteries} presents the accuracy loss and area reduction of the cross-approximate circuits generated by our framework.
The corresponding values are calculated over the respective baseline circuits.
Approximate circuits of very high accuracy (accuracy loss $\leq1\%$ ) are highlighted in green.
For the Molex battery our framework generated approximate circuits with less than 15\% accuracy loss for all the examined models.
Specifically, 19 out of the 24 circuits feature negligible accuracy loss ($\leq1\%$ ).
Similarly, for the Zinergy battery 23 out of the 24 ML circuits can be battery powered while 11 exhibit less than $1\%$ accuracy loss.
Our framework enables battery operation of complex printed ML circuits, while the state-of-the-art only simple ML models~\cite{Mubarik:MICRO:2020:printedml,DATE22:Armen}.
In~\cite{Mubarik:MICRO:2020:printedml} only Decision Trees and SVM-Rs are used, while for example considering the Zinergy battery,~\cite{DATE22:Armen} supports only MLP-Rs with up to 24 coefficients.
On the other hand, our framework can power MLP-Rs with 48 coefficients as well as MLP-Cs.

\begin{figure}[t!]
\includegraphics[]{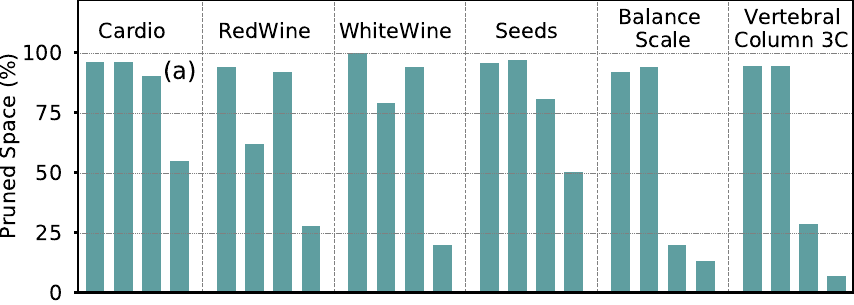} \\
\includegraphics[]{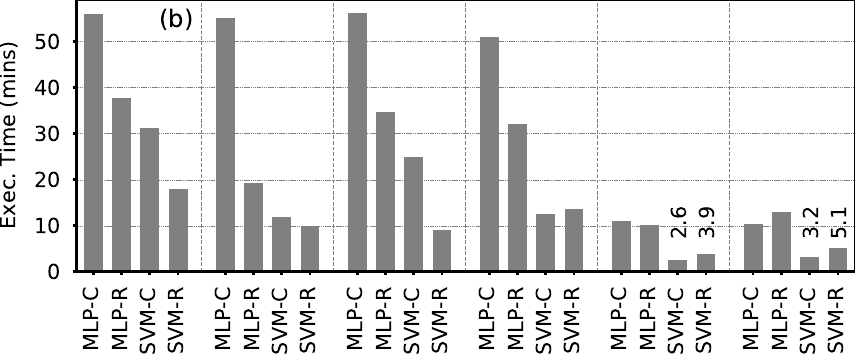}
\caption{
a) Percentage of pruned space achieved by our genetic algorithm's optimizations when evaluating each examined model with 15mW power and 15\% accuracy constraints (see Table~\ref{tab:batteries}).
\yellow{b) Average execution time for 1 epoch of our genetic algorithm for each examined model.}
}
\label{fig:exectime}
\end{figure}

Finally, Fig.~\ref{fig:exectime} evaluates the time complexity of our framework.
First, Fig.~\ref{fig:exectime}a shows the reduction of the design space of tentative solutions implemented in our framework (see Section~\ref{sec:dse}).
To generate Fig.~\ref{fig:exectime}a, the Zinergy 15mW battery is targeted and maximum 15\% accuracy loss is considered.
As shown, the $91\%$ on average of the design space of MLPs is pruned, while in SVMs the corresponding value becomes $48\%$.
As a result, when high design space pruning ratio is achieved, our genetic algorithm can cover most of the space and \yellow{converges promisingly to near-optimal solutions fast-enough (withing only a few epochs).}\label{commentR1C1b} 
It should be noted that, in case of SVM-Rs the design space reduction ranges from 7\% to 55\%.
This is attributed to the fact that the less complex models feature reduced power overheads and thus many cross-approximation solutions can satisfy the given battery constraint.
However, as we will explain later the evaluation of such solutions is very fast (due to the decreased circuit complexity) and thus, we can increase the number of epochs in our optimization without any significant execution time overhead.
Indicatively, the results of Table~\ref{tab:batteries} for the Zinergy 15mW battery are obtained with only 4 epochs for MLPs and 8 epochs for SVMs.
Fig.~\ref{fig:exectime}b presents the \yellow{the average execution time required for one epoch for each examined ML circuit, after repeating each experiment 100 times.
}
The aforementioned execution times refer to a dual-CPU Intel Xeon Gold 6138 server, using only $20$ threads since we are limited by the number of available EDA licenses.
As shown, the average execution time per epoch is only \yellow{$22$} minutes.
The execution time goes up to 57 minutes for several MLP-C that feature however a highly pruned design space.
On the other hand, the SVM-R, that attained low design space reduction, requires only \yellow{10} minutes on average.
\yellow{For some classifiers (e.g., V3 MLP-C and SVM-C) their execution time is lower than that of the respective regressors.
This is explained by the fact that these classifiers are more error resilient compared to the respective regressors and could endure more approximation for the same accuracy loss constraint.
In addition, the respective exact classifiers and regressors feature similar area complexity.
Hence, our genetic algorithm had to explore less complex designs during its optimization phase, leading to reduced simulation times.
}\label{commentR2C1}

\section{Related Work}\label{sec:related}

Recent research around printed electronics at a variety of application domains has been greatly increased.
Applications such as radio-frequency identification (RFID) tags were presented in~\cite{Myny:2021:dualinput}, where a pseudo-CMOS logic for high performance thin-film circuits was designed, while in~\cite{Weller:ASPDAC:2020} a 2-input neuron that can be used to support also a MAC operation was fabricated.
More recently, a flexible 32-bit microprocessor was also fabricated by ARM~\cite{Biggs:Nature2021:flexarm} with over $18,000$ gates.

Due to the inherent ultra resource constrained nature of microprocessors, research on bespoke architectures for reducing area and power by tailoring a processor to an application, is growing exponentially.
Targeting efficient architectures for printed microprocessors,~\cite{Bleier:ISCA:2020:printedmicro} pruned the ISA and the respective microarchitecture accordingly and generated tiny printed microprocessors.
Similarly, \cite{BespokeProcessor} exploited the uniqueness of a given application and proposed a microprocessor IP core, where all logic that will not be used by the application is eliminated.
However, research on printed ML applications is still at its infancy, due to the large feature size of printed circuits.
\cite{Ozer2019Bespoke} proposed an automated methodology to generate bespoke classifiers, while a system integration with hardwired machine learning processor for an odour recognition application was fabricated in~\cite{Ozer:Nature:2020}.
Moreover, \cite{Weller:2021:printed_stoch} employed Stochastic Computing (SC) to reduce area and power of printed MLPs, but at the cost of a prohibitive accuracy loss in most cases.

Targeting to alleviate the increased area and power demands of ML applications, approximate computing has gained a vast research interest.
Significant growth is reported in approximate arithmetic units such as adders and multipliers~\cite{Honglan:JPROC:2020:axsurv}.
Including VOS in a cross-layer approximation, \cite{vader:zerv} proposed an automated synthesis framework for approximate adders and multipliers, while MACACO~\cite{macaco:roy} presented a methodology that can be utilized to analyze how an approximate circuit behaves to timing-induced approximations such as VOS.
Furthermore, \cite{Zervakis2019MultiLevel} exploits the precision-scaling technique that usually leads to lower circuit delay, to perform aggressive voltage scaling for higher impacts on energy savings.
On the other hand, our work distinguishes from state-of-the-art works and proposes \yellow{a design methodology that enables fully-customized voltage-overscaled circuits.
Such a degree of circuit-customization is not available in silicon-based technologies due to the immense associated costs.}\label{commentR1C3a}

\section{Conclusion}
\orange{Printed electronics prevail as a promising solution for a large number of application domains that require ultra low-cost, conformality, low time-to-market, nontoxicity, etc.
Nevertheless, large feature sizes in printed electronics prohibit the implementation of complex circuits.}
To this end, we propose a model-to-circuit cross-approximation built upon bespoke implementations and we demonstrate that our cross-approximation delivers area- and power- efficient approximate printed ML circuits with negligible accuracy loss.
Moreover, we propose a framework to identify, fast-enough, cross-approximated printed ML circuits that satisfy given battery constraints and we show that for 2 out of the 3 examined printed batteries, our framework enables, almost always, battery-operated ML classifiers. 


\bibliographystyle{IEEEtran}
\bibliography{references}

\begin{IEEEbiography}[{\includegraphics[width=1in,height=1.25in,clip,keepaspectratio]{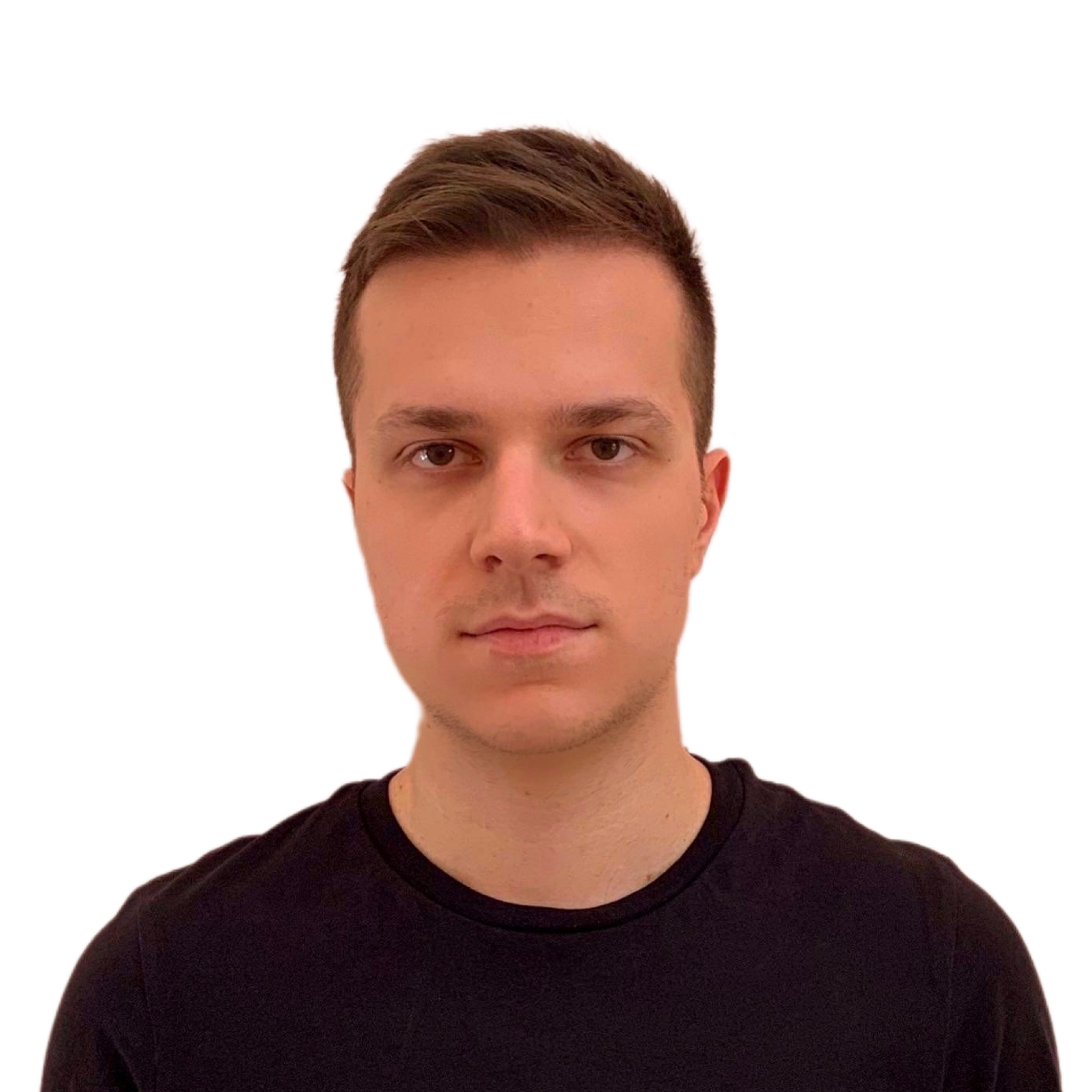}}] {Giorgos Armeniakos} received the Diploma degree from the Department of Electrical and Computer Engineering (ECE), National Technical University of Athens (NTUA), Greece, in 2020, where he is currently pursuing the Ph.D. degree. He holds one best paper nomination at DATE'22 for his work on approximate printed electronics. His research interests include approximate computing, digital circuit design, low power design, machine learning, and optimization.
\end{IEEEbiography}

\begin{IEEEbiography}[{\includegraphics[width=1in,height=1.25in,clip,keepaspectratio]{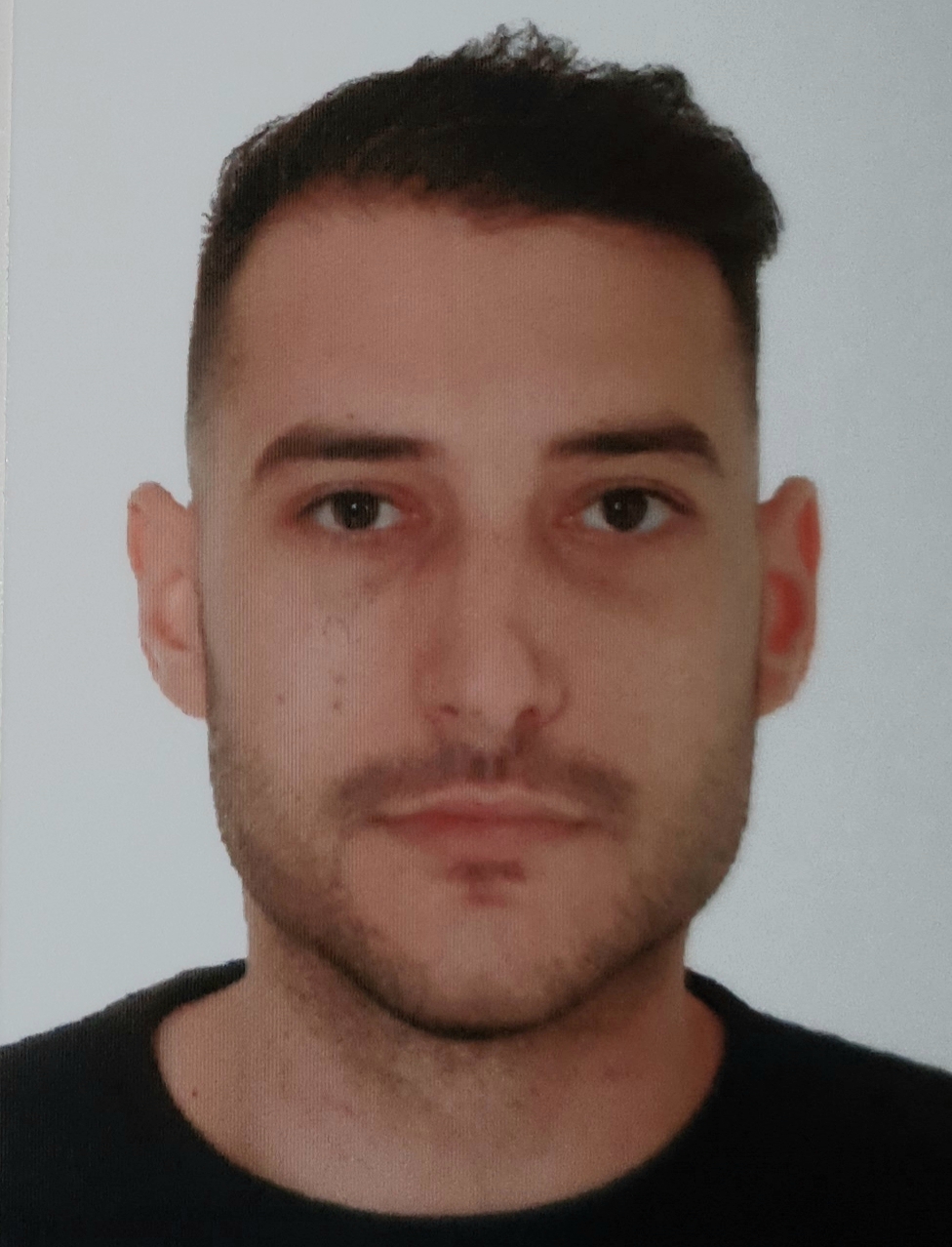}}] {Georgios Zervakis}  is an Assistant Professor at the University of Patras. Before that he was a Research Group Leader at the Chair for Embedded Systems (CES), at the Karlsruhe Institute of Technology (KIT) from 2019 to 2022. He received the Diploma and Ph.D. degrees from the School of Electrical and Computer Engineering (ECE), National Technical University of Athens (NTUA), Greece, in 2012 and 2018, respectively. From 2015 to 2019, Dr. Zervakis worked as a principal investigator in many EU-funded research projects as a member of the Institute of Communication and Computer Systems (ICCS), Athens, Greece. Dr. Zervakis serves as a reviewer in many IEEE and ACM Transactions journals and is also a member of the technical program committee of several major design conferences. He has received one best paper nomination at DATE 2022. His main research interests include low-power design, accelerator microarchitectures, approximate computing, and machine learning.
\end{IEEEbiography}

\begin{IEEEbiography}[{\includegraphics[width=1in,height=1.25in,clip,keepaspectratio]{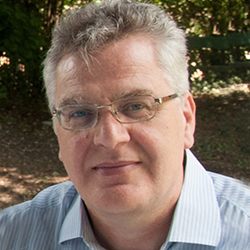}}] {Dimitrios Soudris} (Member, IEEE) received
the Diploma and Ph.D. degrees in electrical engineering from the University of Patras, Patras,
Greece, in 1987 and 1992, respectively. Since
1995, he has been a Professor with the Department of Electrical and Computer Engineering,
Democritus University of Thrace, Xanthi, Greece.
He is currently a Professor with the School of
Electrical and Computer Engineering, National
Technical University of Athens, Athens, Greece.
He has authored or coauthored more than 500 papers in international journals/conferences. He has coauthored/coedited seven Kluwer/Springer books.
He is also the leader and a principal investigator in research projects funded
by Greek Government and Industry, European Commission, ENIAC-JU,
and European Space Agency. His current research interests include high performance computing, embedded systems, reconfigurable architectures,
reliability, and low-power VLSI design. He was a recipient of the award
from INTEL and IBM for EU Project LPGD 25256; ASP-DAC 05 and VLSI
05 awards for EU AMDREL IST-2001-34379, and several HiPEAC awards.
He has served as the general/program chair in several conferences.
\end{IEEEbiography}

\begin{IEEEbiography}[{\includegraphics[width=1in,height=1.25in,clip,keepaspectratio]{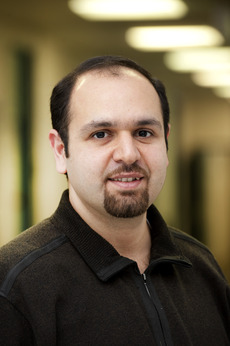}}] {Mehdi B. Tahoori} (M'03, SM'08, F'21) is Professor and the Chair of Dependable Nano-Computing at Karlsruhe Institute of Technology, Germany. He received the B.S. degree in computer engineering from Sharif University of Technology, Iran, in 2000, and the M.S. and Ph.D. degrees in electrical engineering from Stanford University, Stanford, CA, in 2002 and 2003, respectively. He is currently the deputy editor-in-chief of IEEE Design and Test Magazine. He was the editor-in-chief of Microelectronic Reliability journal. He was the program chair of VLSI Test Symposium in (VTS) 2021 and 2018, and General Chair of European Test Symposium (ETS) in 2019. He is the chair of the IEEE European Test Technology Technical Council (eTTTC). Prof. Tahoori was a recipient of the US National Science Foundation Early Faculty Development (CAREER) Award in 2008. He has received a number of best paper nominations and awards at various conferences and journals. He is a recipient of European Research Council (ERC) Advanced Grant.
\end{IEEEbiography}

\begin{IEEEbiography}[{\includegraphics[width=1in,height=1.25in,clip,keepaspectratio]{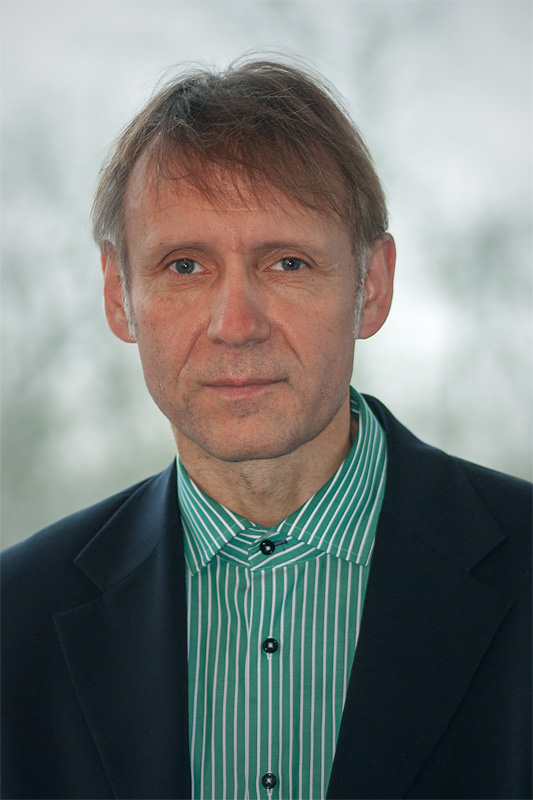}}]{J\"org Henkel} (M'95-SM'01-F'15) is the Chair
Professor for Embedded Systems at Karlsruhe
Institute of Technology. Before that he was a research staff member at NEC Laboratories in
Princeton, NJ.
He has received six best paper awards from, among others, ICCAD, ESWeek and DATE.
For two terms he served as the Editor-in-Chief for the ACM Transactions on Embedded Computing Systems.
He is currently the Editor-in-Chief of the IEEE Design\&Test Magazine and is/has been Associate Editor for major ACM and IEEE Journals.
He has led several conferences as a General Chair incl. ICCAD, ESWeek and serves as Steering Committee chair/member for leading conferences and journals for embedded and cyber-physical systems. Prof. Henkel coordinates the DFG program SPP 1500 ``Dependable Embedded Systems'' and is a site coordinator of the DFG TR89 collaborative research center ``Invasive Computing''. He is the chairman of the IEEE Computer Society, Germany Chapter, and a Fellow of the IEEE.
\end{IEEEbiography}



\end{document}